\documentclass[review]{elsarticle}

\usepackage{lineno}
\usepackage{amsmath}
\usepackage{amssymb}
\usepackage{subfigure}
\usepackage{multirow}
\usepackage[normalem]{ulem}
\usepackage[figuresright]{rotating}
\usepackage[colorlinks,linkcolor=red]{hyperref}
\useunder{\uline}{\ul}{}

\journal{Journal of \LaTeX\ Templates}

\begin{document}

\begin{frontmatter}

\title{A Survey of Robust Adversarial Training in Pattern Recognition: Fundamental, Theory, and Methodologies}

\author[SAT]{Zhuang Qian}

\author[Duke]{Kaizhu Huang\corref{mycorrespondingauthor}}
\cortext[mycorrespondingauthor]{Corresponding author}

\author[SAT]{Qiu-Feng Wang}

\author[undefine1]{Xu-Yao Zhang}


\address[SAT]{School of Advanced Technology of Xi'an Jiaotong-Liverpool University}
\address[Duke]{Institute of Applied Physical Sciences and Engineering, Duke Kunshan University}
\address[undefine1]{Institute of Automation, Chinese Academy of Sciences; School of Artificial Intelligence, University of Chinese Academy of Sciences}

\begin{abstract}
In the last a few decades, deep neural networks  have achieved remarkable success in machine learning, computer vision, and pattern recognition.  Recent studies however show that neural networks (both shallow and deep) may be easily fooled by certain imperceptibly perturbed input samples called adversarial examples.  Such security vulnerability has resulted in a large body of research in recent years because real-world threats could be introduced due to vast applications of neural networks. To address the robustness issue to adversarial examples particularly in pattern recognition, robust adversarial training has become one mainstream. Various ideas, methods, and applications have boomed in the field. Yet, a deep understanding of adversarial training including  characteristics, interpretations, theories, and connections among different models has still remained  elusive. In this paper, we present a comprehensive survey trying to offer a systematic and structured investigation on robust adversarial training in pattern recognition. We start with fundamentals including  definition, notations, and properties of adversarial examples. We then introduce a unified theoretical framework for defending against adversarial samples - robust adversarial training with visualizations and interpretations on why  adversarial training can lead to model robustness. Connections will be also established between adversarial training and other traditional learning theories. After that, we  summarize, review, and discuss various methodologies with adversarial  attack and defense/training algorithms in a structured way. Finally, we present  analysis, outlook, and remarks of adversarial training. 
\end{abstract}

\begin{keyword}
\texttt{adversarial examples\sep adversarial training \sep robust learning }
\end{keyword}

\end{frontmatter}


\section{Introduction}
In the latest decades, the progress of Deep Neural Networks (DNNs) have enabled machine learning to outperform humans in many practical tasks. For example, ResNet-152~\cite{resnet} achieves 3.5\% error rate, better than 5.1\% from human in the challenging ImageNet dataset~\cite{imagenet} for image recognition; Microsoft also designs a system with deep learning that is able to beat human performance in controlled speech recognition~\cite{xiong2016achieving}; LipNet attains  $93.4\%$ accuracy in reading lips from vision data, which is much higher than  professional human lip readers~\cite{lipnet}. 

\begin{figure}[ht]
	\centering
	\includegraphics[width=0.65\textwidth]{{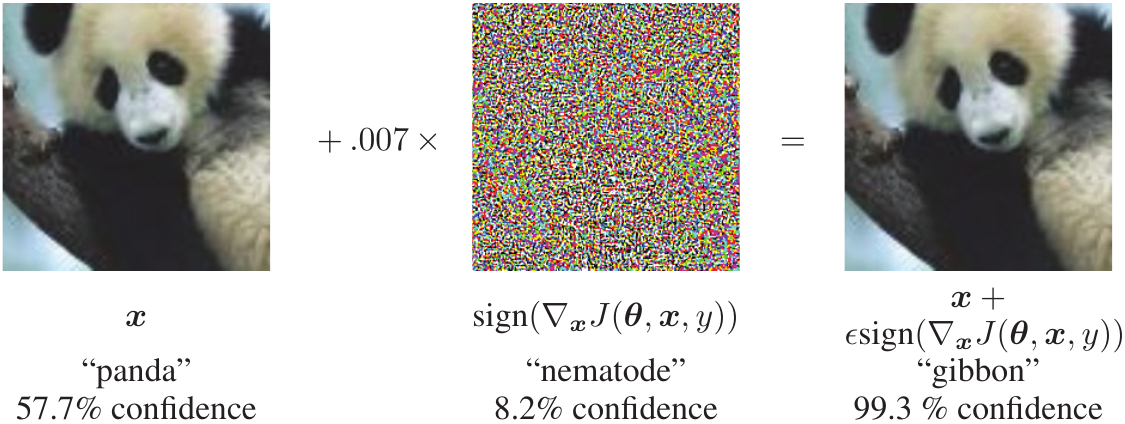}}
	\caption{Adversarial example causes panda pictures to be misclassified as gibbons~\cite{AE}.} 
	\label{fig:AE}
\end{figure}

The success of modern machine learning techniques in performing various complex tasks of pattern recognition has made the security of learning algorithms increasingly important. Robustness, both to accidents and malicious agents, is clearly another vital determinant of the success of machine learning systems in the real world. However, recent studies~\cite{intriguing,AE} have revealed that DNNs (and also shallow NNs) are often vulnerable to the so-called adversarial perturbations that are usually crafted and imperceptible but lead to undesired network outputs. For an illustrative pattern recognition example as shown in Figure~\ref{fig:AE}~\cite{AE},  if one crafts certain adversarial perturbation (the middle image ) on the original input (the left image), the resulting panda image (the right image) called adversarial example will be mis-classified as gibbon by most of the best classifiers with a high confidence~\cite{intriguing,AE}! Apparently, failure to generate a robust DNN would posit huge challenges especially in some security-critic fields such as automatic financial prediction, medical diagnosis, and autonomous driving.

To tackle the security vulnerability arising from real-world pattern recognition tasks, a fast growing body of  proposals have been developed trying to study and promote the model robustness. 
In general, these methods can be categorized into three types: 1) detecting-based methods, 2) denoising-based methods, and 3) robust adversarial training. The detecting-based methods try to discriminate whether or not a sample is an adversarial example. If yes, it will be rejected by the classifier before classification~\cite{hendrycks2016baseline,metzen2017detecting,feinman2017detecting,xu2017feature,pang2017robust}. The denoising-based methods aim to ``purify" adversarial perturbations before the model classifies the images~\cite{wong2018provable, dhillon2018stochastic, xie2017mitigating, meng2017magnet, samangouei2018defense, lamb2018fortified, yang2019me, guo2017countering}. However,these two kinds of methods may be bypassed by  sophisticated attacks (e.g. C\&W~\cite{CW} which we will elaborate later) if the attackers are aware of the defence strategy. On the other hand, the goal of these two types of approaches is not to directly improve the robustness of the model but to make the input less ``deceptive" by auxiliary models or preprocessing. Therefore, with these  methods, it is not easy to analyze the properties of  adversarial samples as well as the robustness of the model itself. In comparison, the third type of methods, i.e. adversarial training, has become the mainstream which is more theoretically founded. By feeding into pattern recognition systems augmented adversarial examples iteratively and stabilizing the loss function in the small neighborhood~\cite{ALP,adversarial_noise_layer,bilateral,fs,zhang2019generalized}, robust adversarial training has been widely studied and applied in the last couple of years.   

In this survey, we will focus on the third family - robust adversarial training and try to present a more dedicated, systematic, and comprehensive survey ranging from  fundamentals, mathematical theory, interpretations, to various methodologies.  
We will elaborate how to perform robust adversarial training so as to  promote the robustness of various pattern recognition algorithms. Different from existing reviews or surveys in adversarial training or adversarial examples~\cite{overview,overview2,overview3}, we  focus on a more systematic way to explain first the fundamentals followed by  a theory which unifies the adversarial training into a general framework; 
visualizations, connections, and  interpretations will also be set out to  gain better understandings of the subject. We will then categorize different methodologies trying to offer an overall picture  in robust adversarial training. Further research issues will  be discussed for the purpose of envisioning current research focus and/or future work in this field.


This work is not only  to provide a review or tutorial for researchers who may just start to explore robust learning, but more importantly to offer a comprehensive picture and a clear guidance towards  up-to-date robust adversarial training methods and research directions. We also expect that it can inspire ideas in common pattern recognition and machine learning, as robust adversarial training can also be readily generalized to other learning  contexts. 



\section{Fundamentals}
In this section, we will introduce fundamentals of robust adversarial training including the definition, notations, categories of adversarial examples as well as their properties.



\subsection{What Are Adversarial Examples?}
\textbf{Definition}. \emph{adversarial example} is informally defined as \emph{an perturbed input sample with imperceptible  noise that would deceive the model to output incorrect results}. As illustrated in Figure~\ref{fig:AE}, the original panda picture can be correctly classified by a DNN with 57.7\% confidence. However, after the crafted imperceptible noise is imposed, the model mis-classifies the picture as a gibbon with 99.3\% confidence, although the picture still looks like a panda to human visually. Such imperceptible noise is often referred to as \emph{adversarial perturbation}.

\begin{figure}[htbp]
	\centering
    \subfigure[A physical perturbation leads that a DNN mis-classifies a Stop sign to a Speed Limit 45 sign~\cite{eykholt2018robust}.]{
		\includegraphics[width=0.35\textwidth]{{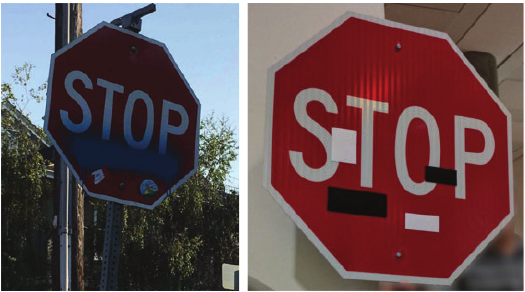}}
		\label{subfig:AE2_1}
		}\qquad
	\subfigure[Perturbing image to impersonate. Crafted glasses lead a DNN to mis-calssify the left to the right actress~\cite{sharif2016accessorize}.]{
		\includegraphics[width=0.4\textwidth]{{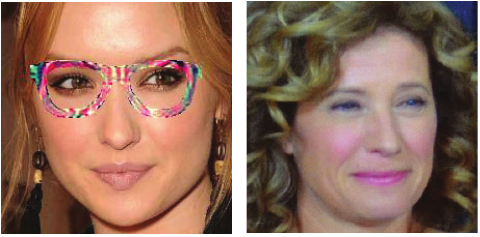}}
		\label{subfig:AE2_2}
	}

	\caption{Some ``inconspicuous" adversarial examples.
	}	\label{fig:AE2}
\end{figure}

It should be noted that adversarial examples could be also visually perceptible but  \emph{inconspicuous}. Such illustrations can be seen in Figure~\ref{fig:AE2}. Clearly, the physical perturbation as imposed in (a) or extra glasses are not imperceptible but ``inconspicuous". These inconspicuous perturbation could also mislead the systems and thus can be considered as adversarial examples. Nonetheless, in this work, we will still focus on imperceptible adversarial examples, which are mathematically easier to be defined. As such, more theoretical analysis and interpretations can be made. 

Last,  adversarial examples have recently been extended to the so-called invariance adversarial examples (IAE) or attacks ~\cite{invariance1,invariance2, invariance3}. Unlike  the above defined adversarial examples,  IAEs aim to change largely the semantics of an image so that the model still leads to the same category output. While IAEs  also present challenges to pattern recognition models, they are actually much different concepts. We will refer the readers to~\cite{invariance1,invariance2, invariance3} for more details.



\textbf{Notations}. Mathematically, a natural input sample and its corresponding ground-truth are typically defined as $x$ and $y$ drawn from data $\mathcal{D}$ respectively. An adversarial example for a specific $x$ is usually defined as $x'$ (or $x_{adv}$). The difference between $x'$ and $x$ defined as $d(x, x')$ is usually constrained within a small distance denoted as $\epsilon$. $d(\cdot)$ can be defined in various norms, e.g. $l_1$, $l_2$ or $l_\infty$ norm. If we denote the pattern recognition model (or a DNN model) by $f$ which is parameterized as $\theta$, an adversarial example seeks to satisfy:
\begin{eqnarray}
\begin{aligned}
d(x, x')  \le \epsilon, \text{such that } \hat{y}(x') \neq \hat{y}(x),
\label{eq:AE1}
\end{aligned}
\end{eqnarray}
where $\epsilon$  is used to bound the magnitude of  adversarial perturbations. $\hat{y}(x)$ denotes the final output of the model $f(x, \theta)$. In this survey, we take the classification model as an example where $\hat{y}(x)=\arg\max f(x, \theta)$ typically. We should however bear in mind that it can be readily generalized to other scenarios like regression without loss of generality. 

To make this problem more tractable, the basic optimization for adversarial examples can be defined as:
\begin{eqnarray}
\begin{aligned}
x_{adv} = \mathop{\arg\max}\limits_{x'} \mathcal{L}(x, \theta) \quad \mathrm{s.t.} \quad \left\| x' - x \right\|_p \le \epsilon,
\label{eq:AE2}
\end{aligned}
\end{eqnarray}
where $\mathcal{L}(\cdot)$ denotes the loss function. The goal of Equation~(\ref{eq:AE2}) is to find the worst sample $x_{adv}$ that maximizes the loss $\mathcal{L}$ of the classifier in a $\ell_p$-ball centered at $x$ with a radius $\epsilon$.

\textbf{Categories of Adversarial Examples}. Adversarial examples can be categorized into white-box and black-box according to the amount of information available to the attacker. In white-box attacks, the attacker is assumed to have access to all information about the target model, including the model architecture, detailed parameters, even training procedure and training data. In black-box attacks, the attacker is assumed to only get access to the output of the target model. More specifically, the black-box attacks can be further divided into restricted black-box attacks (which can only obtain the top few confidence scores, or even only the labels) and unrestricted black-box attacks (which can obtain a full prediction score). Black-box attacks are actually more realistic because attackers usually do not have access to important information (e.g. model architecture and detailed parameters) about the victims in reality. However, white-box attacks are equally importantly, since they are an important approach to measure the performance of a model in the worst-case scenario.

Adversarial examples can also be categorized into targeted or non-targeted attacks according to the purpose of the attack. In targeted attacks, the attackers aim to mislead the adversarial examples as one specified class by the model. While in non-targeted attacks, the attacks are deemed successful as long as  the resulting adversarial examples can  fool the model.

\subsection{Properties of Adversarial Examples}
Following the definition of adversarial examples, three properties of adversarial examples can be summarized. 

\textbf{Imperceptibility}. As mentioned above, antagonistic samples are visually imperceptible in general, meaning that $\epsilon$ is usually small (e.g. 8/255 for images with values from 0 to 255). ``inconspicuous" adversarial examples may be more realistic which are however difficult to be defined mathematically. 

\textbf{Cross Model Generalization}. Transferability~\cite{intriguing,AE,transfer1,transfer2} is another interesting property of adversarial examples, which means an adversarial example generated from one specific model has the ability to misguide other models.


\textbf{Cross Training-set Generalization}.
Well crafted adversarial examples can deceive a classifier that is trained with different data. Such property is called cross training-set generalization. In other words, adversarial examples cannot be defensed by intentionally selecting training samples.  

Adversarial examples with the three properties posit challenges to existing machine learning  algorithms. Next, we will start from a unified theory to explain why adversarial examples are difficult to be defended.

\section{Unified Adversarial Training Theory}
\label{sec:section3}


This section will introduce a unified framework to address adversarial examples and robust adversarial training theoretically. We try to connect adversarial training and gradient regularization based on~\cite{unified} and then visually interpret adversarial examples' transferability. We will also introduce how adversarial training affects the decision boundary and thus leads to model robustness.

\subsection{Adversarial Training and Gradient
Regularization}

\subsubsection{Conventional Training vs.   Adversarial Training}

Without considering any regularization, conventional classification  can be usually formulated as: $\min_{\theta} \mathcal{L}(x, \theta)$, where $\mathcal{L(\cdot)}$ denotes the loss function (e.g. cross-entropy or square error), $x$ denotes the data, and $\theta$ denotes a set of model parameters. Differently, the core of adversarial training is to build a robust model which can generalize well on any samples with small perturbation defined as $\epsilon$, which can be formula a min-max problem as:
\begin{eqnarray}
\begin{aligned}
\min_{\theta} \max_{\epsilon:\parallel \epsilon \parallel_{p} \le \sigma} \mathcal{L}(x+\sigma, \theta).
\label{eq:AT}
\end{aligned}
\end{eqnarray}
The inner maximization tries to find a worst perturbation (i.e. adversarial perturbation) that maximizes the loss for any corrupted data ($x+\epsilon$) constrained by a small magnitude ($\left\| \epsilon\right\|_{p} \le \sigma$); the outer minimization aims to obtain the optimal classifier parametrized by $\theta$ by minimizing the loss function even if the data are corrupted. Once the min-max optimization problem is solved, a classifier $\theta$ can be considered as robust to any imperceptible perturbation over the data, thus leading to a model resisting adversarial examples.  However, Equation~(\ref{eq:AT}) is difficult to be solved due to the non-convex nature of $\sigma$ and $\theta$.

\subsubsection{General Framework}\label{general}
Lyu el al. approximate the inner max problem by its first order Taylor expansion at $x$~\cite{unified}, which leads to the following relaxed formulation:
\begin{eqnarray}
\begin{aligned}
\max_{\epsilon} \mathcal{L}(x) + \nabla_x\mathcal{L}^T\epsilon  \quad \mathrm{s.t.} \quad \left\| \epsilon \right\|_p \le \sigma.
\label{eq:max}
\end{aligned}
\end{eqnarray}
Applying Lagrangian multiplier method, one can obtain a closed-form solution:
\begin{eqnarray}
\begin{aligned}
\epsilon = \sigma~\text{sign} (\nabla\mathcal{L})(\frac{\left| \nabla\mathcal{L}\right|}{\left\| \nabla\mathcal{L}\right\|_{p^*}})^{\frac{1}{p-1}},
\label{eq:epsilon}
\end{aligned}
\end{eqnarray}
where $p^*$ is the dual of $p$, i.e., $\frac{1}{p^*} + \frac{1}{p} = 1$. After substituting the optimal $\epsilon$ back to the original optimization problem (Equation~(\ref{eq:AT})), the influence of perturbations can be formulated as a regularization term named gradient
regularization:
\begin{eqnarray}
\begin{aligned}
\min_{\theta}
\label{eq:grad_reg}\mathcal{L}(x) + \sigma\left\| \nabla_x \mathcal{L}\right\|_{p^*}
\end{aligned}
\end{eqnarray}

\textbf{Remarks}. Recalling the conventional learning paradigm defined as $\min_{\theta}\mathcal{L}(x)$, one can note that the only difference of adversarial training is the additional regularization term. This regularization term is however very special, which is rarely seen in the literature. Specifically, unlike most previous data-independent regularization terms (e.g. 2-norm regularization in support vector machines), the gradient regularization term is data-specific/dependent, meaning that it is related to each data $x$. Interestingly, if we design a special loss function as a linear form e.g.  $\mathcal{L}(x)=|\theta^Tx-y|$, feeding adversarial examples implies a 1-norm or 2-norm regularization  during training when $p=\infty$ or $p=2$ respectively. This intuitively shows that traditional 2-norm or 1-norm regularization used to promote the model robustness is a special case of adversarial training.

\subsubsection{Special Cases}\label{specialcase}

\textbf{Case $p=\infty$}.  Assuming $\left| \nabla\mathcal{L}\right| > 0$, the worst-case perturbation $\epsilon$ becomes $\epsilon = \sigma~\text{sign} (\nabla\mathcal{L})$.
In this case, the general method is reduced to Fast Gradient Sign Method (FGSM)~\cite{AE}. If the worst case perturbation is performed multiple steps, it becomes the famous PGD~\cite{PGD} attack ( to be introduced shortly). Meanwhile, the corresponding induced gradient regularization term is $\sigma\left\| \nabla_x \mathcal{L}\right\|_{p^*} = \sigma\left\| \nabla_x \mathcal{L}\right\|_{1}$.
In this case, the gradient is not penalized in an isotropic way, which may decrease the performance, especially when the data are preprocessed to be Gaussian-like~\cite{unified}.

\textbf{Case $p=2$}. Lyu et al.~\cite{unified} take the case $p=2$ as the \emph{standard gradient regularization}, where the worst-case perturbation $\epsilon$ can be formulated as $\epsilon = \sigma\frac{\nabla\mathcal{L}}{\left\| \nabla \mathcal{L}\right\|_{2}}$.
The standard gradient regularization induced first order regularizer becomes $\sigma\left\| \nabla_x \mathcal{L}\right\|_{p^*} = \sigma\left\| \nabla_x \mathcal{L}\right\|_{2}$.
In this case, the  standard gradient regularization is very similar to the Jacobian Regularization~\cite{jacb}, describing how the output is changed against the input.

\textbf{Case $p=1$}. When $p=1$, the worst perturbation $\epsilon$ becomes active in one single dimension: $\epsilon_j =\sigma$ if $\nabla_x\mathcal{L}_j = \max_j \mathcal{L}_j$; $\epsilon_j=0$ otherwise. 
The induced gradient regularization term is $\sigma\left\| \nabla_x \mathcal{L}\right\|_{p^*} = \sigma\left\| \nabla_x \mathcal{L}\right\|_{\infty}.$
Lyu et al.~\cite{unified} argue that in this case, all perturbations are put into one direction, and the regularization term is not very appealing since it only penalizes gradient in one direction.

\subsection{Why Adversarial Perturbation Generalizes?}
\label{sec:visualization}
\begin{figure}[htbp]
	\centering
    \subfigure[Original input mages from MNIST dataset.]{
		\includegraphics[width=0.7\textwidth]{{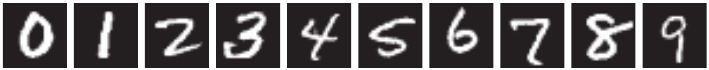}}
		\label{subfig:perturbation1}
		}
	\subfigure[Inputs perturbed with $p=2$ and $\sigma=1$]{
		\includegraphics[width=0.7\textwidth]{{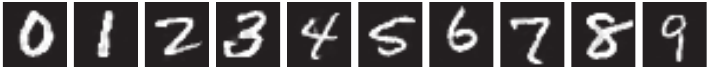}}
		\label{subfig:perturbation2}
	}
	\subfigure[The corresponding perturbations magnified by a factor of 10]{
		\includegraphics[width=0.7\textwidth]{{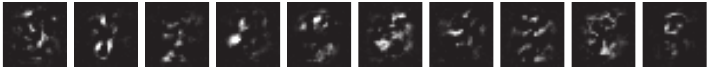}}
		\label{subfig:perturbation3}
	}
	\subfigure[Inputs perturbed  with $p=2$ and $\sigma=10$]{
		\includegraphics[width=0.7\textwidth]{{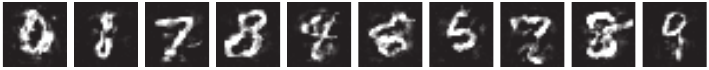}}
		\label{subfig:perturbation4}
	}
	\caption{Visualization of adversarial example in case $p=2$~\cite{unified}.}
		\label{fig:perturbation}
\end{figure}

Taking  $p=2$ as one example, one can easily calculate the adversarial perturbation given a specific sample. With this theory, Lyu et al. have conducted one excellent illustration on MNIST data~\cite{MNIST} as shown in Figure~\ref{fig:perturbation}. 
As observed, 10 slightly perturbed examples in (b) are almost indistinguishable from the original ones in (a). However, if perturbations are magnified by a factor of 10, one can easily spot that  `2' is changed to `7', `3' is changed to `8', and `5' is changed to `8'! In other words, adversarial perturbations tend to identify the direction which  easily morphs the input sample into another similar different class. While imperceptible perturbations may make no sense to human, they could be very sensitive to pattern recognition models. This visualization explains why adversarial examples have cross-model generalization: they may not be just noises to the input in response to the model, but closer samples of the other classes.



\subsection{Why Adversarial Training Promotes Robustness?}
\begin{figure}[htbp]
	\centering
	\includegraphics[width=0.7\textwidth]{{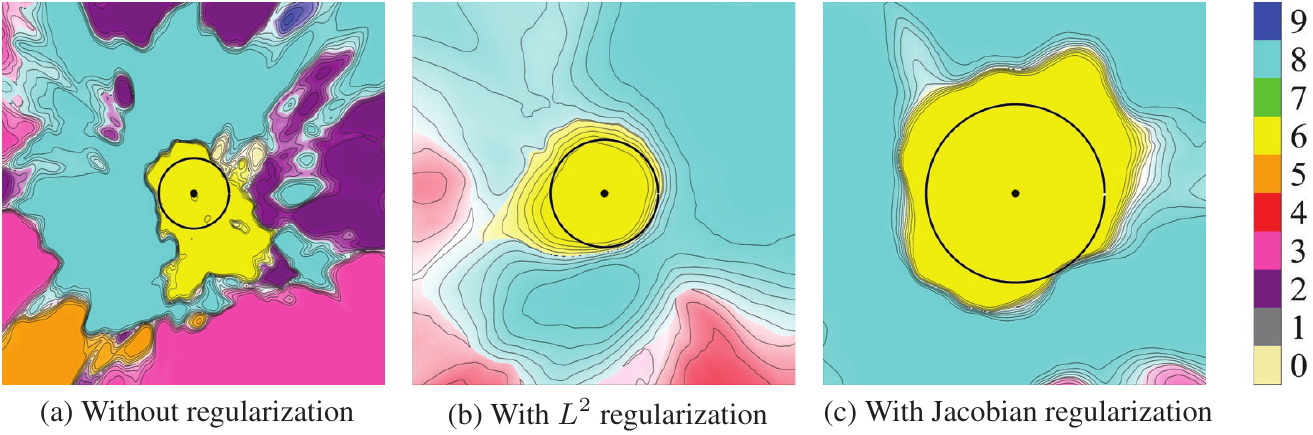}}
	\caption{Effect of different regularizations on the decision boundary (on MNIST)~\cite{jacb}.} 
	\label{fig:jacb1}
\end{figure}

Hoffman et al.~\cite{jacb} propose a computationally efficient Jacobian regularization that increases classification margins of DNNs, which leads to significant improvements in robustness. As mentioned in Section~\ref{specialcase}, the Jacobian regularization is similar to gradient regularization in case $p=2$. Thus we borrow the nice visualization from~\cite{jacb} to illustrate the effect of adversarial training. From Figure~\ref{fig:jacb1}, training with $l_2$ regularization could lead
to a moderately smoother decision boundary, and does not significantly enlarge the margin. This is understandable, as $l_2$ regularization is performed independent of each specific sample as seen in the remarks of Section~\ref{general}. However, training with Jacobian regularization (or adversarial training) which is data specific, the margin around each sample is significantly increased, and the decision boundary is much smoother, as observed in Figure~\ref{fig:jacb1}(c). Namely, after adversarial training is applied, even if the adversarial perturbation is added, the sample will be less likely to be changed into the region of the other class, thereby leading to model robustness. 

\section{Methodologies}
We will introduce different methodologies on adversarial attacks first. We then elaborate robust adversarial training (defense) methodologies. since attack and defence are sometimes difficult to separate, some methods may appear in both parts. Additionally, we also discuss some recent sophisticated  attacks. 


\subsection{Adversarial Attacks}
We briefly review  mainstream adversarial attacks  including white-box and black-box attacks. We do not categorize them by target and non-target attacks as done in some other reviews~\cite{overview}, since the vast majority of approaches can easily be implemented on both the target and non-target settings. 

\subsubsection{White-box Adversarial Attacks}
White-box attacks are essential for evaluating the model robustness. In general, the accuracy of the model on white-box attacked samples from the test set is referred to as the \emph{robust accuracy}. Such model robustness should be reported by the robust accuracy under different methods and different magnitude of white-box attacks . Methods of white-box attacks can be further divided into three categories: 1)
\emph{Optimization-based}:  As shown in Equation~(\ref{eq:AE2}), solving adversarial examples is equivalent to finding the worst perturbation that maximizes the loss function in a neighbourhood centred at the original sample. In summary, this kind of approaches generate adversarial samples by maximizing the classification risk. However, due to the non-linearity of many pattern recognition models, it is often difficult to solve exactly  the optimization problem. 2) \emph{Gradient-based methods:} To generate adversarial examples, the model loss function  can be approximated as a linear function in a very small neighbourhood. Thus, the direction of the worst perturbation is the direction of the gradient of the loss function with respect to the input. 3) \emph{Approximation-based methods}: Some researchers argue that the gradient-based methods cannot accurately estimate the worst sample because the loss function is not necessarily linear in the neighbourhood of some samples. To solve this problem, researchers have proposed the \emph{approximation-based methods}. Specifically, smaller updates are used to approximate the adversarial sample via multiple steps. While these methods are summarized in Table~\ref{tab:attack}, we will briefly discuss 
some representative works.

\noindent \textbf{\emph{$\bullet$ Optimization-based Methods}}

\textbf{L-BFGS Attack}. L-BFGS~\cite{intriguing} is extended from an early-proposed optimization-based attack approach, which aims to solve the  optimization problem as $\mathop{\arg\min}\limits_{r} \left\| r\right\|_2$ where $f(x+r) = t$ and $x+r\in \left[0, 1 \right]$.
Here $r$ defines the adversarial perturbation, $t$ is the target mis-classification label. However, this problem is hard to solve.  Szegedy et al.~\cite{intriguing} approximate it by using a box-constrained Limited Memory Broyden-Fletcher-Goldfarb-Shanno (L-BFGS)~\cite{liu1989limited}:  $\mathop{\arg\min}\limits_{r} c\left\| r\right\|_2 + \mathcal{L}(x+r, t)$,
where  $x+r$ is normalized to [0,1], $\mathcal{L}$ represents the loss function. Since this objective function does not guarantee that $x+r$ is adversarial for $c>0$, $c$ will be iteratively increased until one adversarial sample is found.

\textbf{AMDR}. Adversarial Manipulation of Deep Representations (AMDR)~\cite{AMDR} is proposed to minimize the distance between the latent feature of the original image and the target label image:
\begin{eqnarray}
\begin{aligned}
\arg\min_{x'} \left\| f_l(x') - f_l(x_t) \right\|^2_2,~\text{such that}~ \left\| f_l(x') - f_l(x_t) \right\|_\infty < \epsilon,
\label{eq:AMDR}
\end{aligned}
\end{eqnarray}
where  $f_l(\cdot)$ represents the output of $l$-th layer of a DNN, $x_t$ represents the image from target label $t$.

\textbf{C\&W}. Carlini \& Wagner Attacks (C\&W or CW)~\cite{CW}  approximates the optimization problem similar to L-BFGS by an empirically-chosen loss function:  $\mathcal{L}_{CW}(x', t) = \max \left(\max_{i\neq t} \left\{ Z(x')_{(i)}\right\} - Z(x')_{(t)}, -k  \right)$
where $Z(x')_{(i)}$ denotes the $i$-th component of the classifier’s logits, $t$ denotes the target label, and $k$ represents the minimum desired confidence margin for adversarial examples. Minimizing this equation means that the distance between logit values of class $t$ and the second most-likely class will be minimized until  $t$ becomes the highest logit value and exceeds the second most-likely by $k$. Under $L_\infty$, it can be formulated as:
\begin{eqnarray}
\begin{aligned}
\arg\min_r \left( c \cdot \mathcal{L}(x+r, t) + \sum_i\max(0, r_{(i)}-\tau)  \right)
\label{eq:CW2}
\end{aligned}
\end{eqnarray}
$\tau$ is initialized to 1, and reduced by a factor of 0.9 per iteration  if $r_{(i)} < \tau$, until no adversarial example is found.

\textbf{DeepFool}. DeepFool~\cite{deepfool} estimates the distance between the input and the closest decision boundary. Adversarial examples generated by DeepFool and the minimal estimated distance are both critical measurements of the robustness of the model.

\noindent \textbf{$\bullet$ \emph{Gradient-based Methods}}

\textbf{FGSM}. Fast Gradient Sign Method (FGSM)~\cite{AE} is a classic gradient-based method to  find perturbations direction quickly. As it is one special case of $p=\infty$ of the unified theory in Section~\ref{specialcase}. We will omit its details.



\textbf{BIM}. Basic Iterative Method (BIM)~\cite{BIM, BIM2} is an iterative extension of FGSM~\cite{intriguing} and is
sometimes referred to as Iterative FGSM or I-FGSM~\cite{overview}. BIM iterates FGSM multiple times within a supremumnorm bound on the total perturbation, which can be formulated as 
\begin{eqnarray}
\begin{aligned}
x'_{i+1} = \text{Clip}_{\epsilon}\left\{x'_i + \alpha \cdot \text{sign}(\nabla_x\mathcal{L}(x'_i, y)) \right\},~\text{for}~i = 1 ~\text{to}~n,~x'_0 = x.
\label{eq:BIM}
\end{aligned}
\end{eqnarray}
$n$ is the total iterates number, and $\alpha$ is the step size to the control the magnitude of each step ($\alpha \ge \frac{\epsilon}{n}$ usually). $\text{Clip}_{\epsilon}$ is the clipping operator is defined as $\text{Clip}_{\epsilon} = \min\left\{255, x + \epsilon, \max\left\{ 0, x-\epsilon,x'_i \right\} \right\}$ that constrains each input feature (e.g., pixel) to be within an $\epsilon$-neighborhood of the original samples, as well as within the feasible input space (e.g. [0, 255] for images).  

\textbf{MI-FGSM}. Momentum Iterative FGSM (MI-FGSM)~\cite{MBIM} is proposed to use the momentum to stabilize the directions for perturbations, and also helps to escape from the local optimum:
\begin{eqnarray}
\begin{aligned}
x'_{i+1} = \text{Clip}_{\epsilon}\left\{x'_i + \alpha \cdot \text{sign}(\mu \cdot g_i + \frac{\nabla_x'\mathcal{L}(x'_i, y)}{\left\| \nabla_x'\mathcal{L}(x'_i, y)   \right\|_1}) \right\},~g_0=0,~x'_0 = x.
\label{eq:MBIM}
\end{aligned}
\end{eqnarray}

\textbf{R+FGSM}.Randomized Single-step Attack FGSM  (R+FGSM)~\cite{R+FGSM} improves FGSM by adding a small random perturbation to the input in the beginning. R+FGSM is not usually used only for attacks or robustness evaluation, but a defensive strategy in  improving  FGSM to escape from local optimum.

\textbf{PGD}. Projected Gradient Descent (PGD)~\cite{PGD} is a combination of R+FGSM~\cite{R+FGSM} and BIM~\cite{BIM,BIM2}.
Like R+FGSM, PGD is often used as a defensive strategy. However, PGD is indeed widely used to evaluate model robustness.

\noindent  \textbf{$\bullet$ \emph{Approximation-based Methods}}

\textbf{BPDA}. Backward Pass Differentiable Approximation (BPDA)~\cite{BPDA} is proposed to alleviate the obfuscated gradient problem, i.e., 
 the robustness of some models are overestimated. The reason is that some non-differentiable parts are induced, e.g., due to some non-differentiable preprocessing functions or non-differentiable network layers.   BPDA aims to find a similar and differentiable function to replace the  non-differentiable one  during back-propagation. It is argued that even the gradients are slightly inaccurate after replacement, they are still useful in constructing an adversarial example. BPDA however requires more iterations of gradient descent than without, and applying replacement on both the forward and backward pass might be less/completely ineffective.

\textbf{SPSA}. Simultaneous Perturbation Stochastic Approximation (SPSA)~\cite{SPSA} is also a gradient estimation method estimating the gradient by two queries of the model as $\frac{\partial f}{\partial x} \approx \frac{f(x+\Delta v) - f(x-\Delta v)}{2\Delta} \cdot v,$
where $\Delta$ is a very small constant (e.g. $10^-9$). $v\sim \left\{-1,1 \right\}^D$ is  sampled from the Rademacher distribution whose elements are  either $+1$ or $-1$. SPSA can be used under the black-box setting.

 \textbf{Other Approximation Methods}. Different from numerical gradient estimators like BPDA and SPSA, many approaches generate adversarial examples with an additional generative model~\cite{kos2018adversarial,xiao2018spatially, NAA, ATN,advgan}. For example, Adversarial Transformation Network (ATN)~\cite{ATN} tries to generate adversarial perturbation given an input by a neural network. AdvGAN~\cite{advgan} aims to generate adversarial perturbations by using
a Generative Adversarial Network (GAN)~\cite{GAN}.

\subsubsection{Black-box Adversarial Attacks}
Essentially, black-box attacks estimate the gradient of the target model by various methods (e.g. model-based or numerical gradient estimators). Adversarial samples will be generated accordingly with the estimated gradient.

\textbf{SBA}. Substitute Blackbox Attack (SBA)~\cite{SBA} is an early-proposed method to generate black-box adversarial examples. The fundamental idea is to imitate the target model by training a substitute trained on a synthetic dataset labeled by the target model. It then generates adversarial examples on white-box substitute model to attack the target model. SBA leverages the transferability property of adversarial examples as mentioned in Section~\ref{sec:visualization}.

\textbf{ZOO}. Zeroth Order Optimization (ZOO)~\cite{ZOO} approximates the gradients of the objective function using finite-difference numerical estimates similar to SPSA~\cite{SPSA}. There are two versions: ZOO-Adam and ZOO-Newton. ZOO-Adam applies symmetric difference quotient to estimate the derivatives as $\widehat{g_i}:= \frac{\partial f}{\partial x_i} \approx \frac{f(x+h e_i) - f(x-he_i)}{2h}$
where $h$ is a small constant, $e_i$ is a standard basis vector with only the $i$-th component as 1 otherwise 0.  Adam~\cite{Adam} is then used to optimize the input. ZOO-Newton further adopts a second-order derivative approximation. Thus an estimate of Hessian on the coordinates is required as $\widehat{h_i}:= \frac{\partial^2 f}{\partial x^2_i} \approx \frac{f(x+h e_i) - 2f(x) + f(x-he_i)}{h^2}$.
Unlike the other methods, ZOO adopts stochastic coordinate descent~\cite{CDA} rather than stochastic gradient descent~\cite{SGD}.

\textbf{OPA}. One-pixel Attack (OPA)~\cite{OPA} deceives the target model by modifying a limited number of features (pixels) of the input. OPA aims to find the perturbation from $P$ randomly initialized  candidates during $I$ generation~\cite{overview}:
\begin{eqnarray}
\begin{aligned}
r^{i+1}_p = \arg\max \left(  F(x\bigoplus r^i_p), F(x\bigoplus r')  \right), ~\text{where}~ r'=r^i_{p1} + \lambda(r^i_{p2} - r^i_{p3}).
\label{eq:OPA}
\end{aligned}
\end{eqnarray}
$i$ means the $i$-th generation and $p$ means the $p$-th candidates. $(x\bigoplus r)$ represents modifying the $r$ candidates onto $x$. In the target attack case, $F(\cdot)$ is the prediction probability of model $f$ for a target class $t$ (i.e. $F(\cdot) = f(x \bigoplus r)_{(t)}$). In non-target attack case, $F(\cdot)$ is the negative of probability for the ground-truth class (i.e. $F(\cdot) = -f(x \bigoplus r)_{(y)}$).

\textbf{BA}. Boundary Attack (BA)~\cite{boundary_attack} is also a substitute model-free method to generate black-box attacks. Starting from an existing adversarial sample, BA performs a random walk along the boundary between the adversarial and non-adversarial regions. As such, it stays in the adversarial regions while the distance to the original image is reduced. Finally it can find the minimum distance from the original image while staying in the adversarial regions within $k$ steps.

\textbf{NAA}. Natural Adversarial Attack (NAA)~\cite{NAA} generates natural adversarial examples whose syntax and semantics are coherent. NAA adopts two networks with opposite functions $G$ and $I$.  $G$ is a pre-trained Wasserstein GAN framework~\cite{wassersteinGAN}  that maps a latent representation $z$ onto an instance $x$, while $I$ is a generative model learning to map images $x$ to latent representation $z$. Then an iterative stochastic search or hybrid shrinking search is used to find a mis-classified latent instance  $z'$.

\subsubsection{Recent Sophisticated Attacks}
As mentioned in BPDA~\cite{BPDA}, the robustness of some methods can be overestimated due to obfuscated gradients or gradient masks. Unlike DBPA requiring to replace the non-differentiable part, some recent methods can also generate more sophisticated adversarial samples than PGD and C\&W.

\textbf{AA}. AutoAttack (AA)~\cite{AA} is one ensemble method, which combines four methods. Two of them are proposed by AA itself: Auto-PGD (APGD) (selecting automatically the step size), and Auto-PGD with Difference of Logits Ratio Loss ($\text{APGD}_\text{DLR}$) (replacing  cross-entropy with Difference of Logits Ratio Loss, similar to C\&W). The other two existing complementary methods are FAB~\cite{croce2020minimally} and Square Attack~\cite{andriushchenko2020square}. FAB minimizes the norm of the perturbation necessary to achieve a misclassification, which appears effective on models affected by gradient masking. Square Attack~\cite{andriushchenko2020square} is a score-based black-box attack for norm bounded perturbations by leveraging random search.

\textbf{Rays}. Ray Searching Attack (Rays)~\cite{Rays} is a hard-label adversarial attack method in which  the attacker can only access the prediction (not the predicted probability) of the model. It leverages binary search to locate the decision boundary radius. For example, by initializing direction vector ${+1,+1, ... , +1}$, it conducts a binary search for the first time to project onto the decision boundary, records the magnitude of perturbation at this time, and then randomly modifies the sign bits of the vector, equivalent to turn to another direction. If not in the adversarial region, it then discards and randomly modifies again; otherwise, it can find a smaller perturbation and run the  next binary search.


\subsection{Methodologies of Adversarial Training}
\label{sec:AT}
As discussed in Section~\ref{sec:section3}, adversarial training is one most direct and effective way to defend against adversarial samples. The relationship between adversarial training and adversarial attack can be analogous to  a shield and a spear: to get the most robust model, one need be able to defend against the strongest attack, and vice versa. Therefore, the basic idea of adversarial training is to use the worst-case adversarial samples, i.e., the strongest adversarial samples, to train the model. In the following, we will introduce some most representative methods while leaving a relatively comprehensive  summarization in Table~\ref{tab:defence}. 

\textbf{FGSM-AT}: FGSM adversarial training (FGSM-AT)~\cite{AE} is one of the earliest adversarial training method that trains the model by adding into the training set  FGSM~\cite{AE} adversarial examples iteratively. Namely, the models are required to handle both clean samples and auxiliary adversarial examples $\alpha\mathcal{L}(x, y) + (1 - \alpha)\mathcal{L}(x', y)~~\text{s.t.}~~ x' = x + \epsilon \cdot \text{sign}(\nabla_x\mathcal{L}(x, y))$,
where $\alpha$ is a hyperparameter to balance the weight of the loss terms between normal and adversarial inputs. 
 Although FGSM-AT demonstrates its effectiveness for defending against single-step attacks, even on large-scale datasets~\cite{FGSM_imagenet}. Models trained with single-step attack are found still vulnerable to iterative multi-step generated attacks such BIM and PGD. On the other hand, the so-called \emph{label leaking}~\cite{FGSM_imagenet} may occur. Concretely, Kurakin et al.~\cite{FGSM_imagenet} observe that models trained on FGSM would obtain higher accuracy against FGSM adversarial samples than clean samples during evaluation. They argue that this effect does not only occur in FGSM-AT, but also in other one-step methods leveraging the true label. One-step methods perform a very simple and predictable transformation leveraging the true label,   making the model able to learn to recognize and lead to label leaking. To alleviate this issue, some other one-step methods which  do not directly access the label are used to evaluate the robustness to adversarial examples.

\textbf{PGD-AT}. As a representative work in adversarial training, the PGD-AT~\cite{PGD} has a profound impact. Normally,  PGD-AT~\cite{PGD} is almost evaluated in all the adversarial training literature. Similar to Equation~(\ref{eq:AT}), a variant of adversarial training under a worst-case scenario $\min_{\theta}\mathbb{E}[ \max_{\left\| x'-x\right\|_{\infty}} \mathcal{L}(x', y) ]$ is proposed~\cite{PGD}.
Adversarial examples $x'$ are generated by BIM~\cite{BIM} adding a random initialization of noise at the first iteration:
\begin{eqnarray}
\begin{aligned}
x^{t+1} = \prod_{x+S}(x^t + \alpha\cdot\text{sign}(\nabla_x\mathcal{L}(x, y))),~~\text{where}~~ x^0 = x + U(-\epsilon, \epsilon).
\label{eq:PGD2}
\end{aligned}
\end{eqnarray}
$\prod_{x+S}$ is the projection constraining that  $x'$ is bound by the $l_\infty$ ball $S$ with radius $\epsilon$. $U(-\epsilon, \epsilon)$ means a uniform distribution on $[-\epsilon, \epsilon]$. Unlike the strategy of using both the adversarial and original samples for training in FGSM-AT, PGD-AT uses only adversarial samples in training. It is proved that the adversarial loss arrives at a plateau after several iterations (though starting randomly). 

Moreover, the final loss values of PGD-AT can be shown significantly smaller than on their standard counterparts. However, this is not really a good sign, as it relates to \emph{robust generalization}, which we will introduce later. Although PGD-AT demonstrates its effectiveness against adversarial examples, it substantially reduces the accuracy for clean samples. On the other hand, the computational cost of PGD-AT far exceeds that of standard training, since  multiple back-propagations are needed to obtain gradients with respect to inputs. 

\textbf{Variants of PGD-AT}.
Many researchers have proposed various methods aiming to improve  PGD-AT. For example, Triplet Loss Adversarial (TLA) training~\cite{TLA} leverages the triplet loss to reduce the distance between adversarial and clean samples with similar classes and push away samples of different classes. Adversarial Noise Layer (ANL)~\cite{adversarial_noise_layer} injects adversarial perturbations into latent feature. Bilateral Adversarial Training (BAT)~\cite{bilateral} is designed to perturb both the image and the label during training. Many other methods are also proposed to accelerate the training of PGD-AT, such as Free~\cite{free}, SLAT~\cite{park2021reliably}, GradAlign~\cite{andriushchenko2020understanding}, and Fast AT~\cite{wong2020fast}.

\textbf{Ensemble Adversarial Training}. Tram$\grave{e}$r et al.~\cite{ensemble} argue that single-step adversarial training methods would converge to a degenerated global minimum. Thus the generated adversarial examples are weak to support a robust model. As a result, single-step adversarial training models remain vulnerable to black-box attacks. Ensemble Adversarial Training (EAT) is then proposed which trains the robust model with adversarial examples generated to attack various other \emph{static pre-trained} models. It is argued that perturbations crafted on an external model are good approximations for maximizing the adversarial training objective (e.g. that of PGD). 

Pang et al.~\cite{ensemble_diversity} ensemble several individual adversarial trained classifiers to improve robustness with  Promoting Ensemble Diversity (PED). PED aims to force the maximal prediction to be consistent with the true label, and the other non-maximal predictions to be as diverse as possible. 

\textbf{Adversarial Logit Pairing}. Logit pairing  is proposed to force the logits for pairs of examples to be similar~\cite{ALP}. Two types of logits pairing are designed: Adversarial Logit Pairing (ALP) and Clean Logit Pairing (CLP). ALP aims to align the output of a normal sample with its corresponding adversarial sample after passing through the model; CLP forces the output of any two inputs to be similar after passing through the model. Surprisingly, although the two inputs in CLP are often from different classes, it can significantly improve the robustness of the model. Combining ALP with PGD-AT can substantially improve the robustness of the model to both white-box and black-box attacks.

\textbf{TRADES}.  To trade-off accuracy and robustness, 
the prediction error for adversarial examples (robust error) is decomposed into the natural (classification) error and boundary error~\cite{trades}. TRADES is then proposed to trade adversarial robustness off against accuracy based on their theoretical analysis:
\begin{eqnarray}
\begin{aligned}
\min_f\mathbb{E}\left\{ \underbrace{\phi \left ( f(x)y \right )}_{\text{for accuracy}} +\underbrace{\max_{x'\in\mathbb{B}(x,\epsilon)}\phi \left ( f(x)f(x')/\lambda  \right ) }_{\text{regularization for robustness}}\right\}
\label{eq:trades}
\end{aligned}
\end{eqnarray}
where $(x,y)$ is the data and label respectively, $\phi$ means the classification-calibrated loss, $\mathbb{B}(x, \epsilon)$ represents a neighborhood of $x$ within $\epsilon$, $f(x)$ is the output vector of the learning model.
Replacing $\phi$ with a multi-class calibrated loss $\mathcal{L}(\cdot,\cdot)$, Equation~(\ref{eq:trades}) is transformed into multi-class classifications: $$\min_f\mathbb{E}\left\{\mathcal{L} \left ( f(x), y \right ) +\max_{x'\in\mathbb{B}(x,\epsilon)} \mathcal{L} \left ( f(x), f(x')/\lambda  \right ) \right\}.$$

\textbf{FAT}. Zhang et al.~\cite{fat} argue that adversarial examples generated by PGD~\cite{PGD} sometimes hurt the natural generalization. After adversarial training, the classification boundary is gradually blurred as the number of attack steps gradually increases.  The overpowering attacks like large step PGD  will ``kill" the training process, since the model after adversarial training cannot correctly classify some original samples if it is to correctly classify the adversarial sample. Thus a more robust model will mis-classify normal samples.
To tackle this problem, Friendly Adversarial Training (FAT) is proposed to find ``friendly adversarial samples" that enable confident mis-classification and do not cross the decision boundary too much for training:
\begin{eqnarray}
\begin{aligned}
x'_i = \arg\min_{x'\in\mathbb{B}_{\epsilon}(x_i)}\ell(f(x'), y_i)~~\text{s.t.}~~ \ell(f(x'), y_i) - \min_{y\in Y}\ell(f(x'), y) > \rho
\label{eq:fat}
\end{aligned}
\end{eqnarray}
where $y_i \neq \arg\min_{y\in Y}\ell(f(x'), y)$ ensures that $x'$ is misclassified. The wrong prediction for $x'$ is better than the desired prediction $y_i$ by at least $\rho$ in the loss value. 

\textbf{FS}. As conventional supervised attacks such as PGD push the samples towards the decision boundary, they may ignore the original data manifold structure. To this end, Feature Scattering based Adversarial Training (FS)~\cite{fs} is developed to generate adversarial examples for training through feature scattering in the latent space in an unsupervised fashion. The core of FS is to perturb the local neighbourhood structure by maximizing the optimal transport distance between natural and perturbed examples as $c(x_i,x'_j)=1-\frac{f_\theta(x_i)^\top f_\theta(x'_j) }{\left\|f_\theta(x_i)\right\|_2\left\|f_\theta(x'_j)\right\|_2}.$
FS achieves very encouraging results in defending against FGSM~\cite{AE}, PGD~\cite{PGD} and C\&W~\cite{CW}. However, later works indicate that the performance of FS degrades substantially in defending against more sophisticated attacks~\cite{AA,Rays}.

\textbf{MAT}. Virtual Adversarial Training (VAT)~\cite{virtual_adversarial_training} is proposed to consider local distributional smoothness by training models with virtual adversarial examples which are generated from unlabeled data. Similar to VAT, Manifold Adversarial Training (MAT)~\cite{zhang2021manifold} also considers the local manifold of latent and output representations as a regularization. A latent feature space is derived with Gaussian Mixture Model (GMM) in DNN. Smoothness is defined as the largest variation of Gaussian mixtures given a local perturbation:
\begin{eqnarray}
\begin{aligned}
\max_{\theta}\frac{1}{N}\sum^N_{i=1} \log P(y_i|x_i,\theta)+\lambda_1\frac{1}{N}\sum^N_{i=1}\min_{\epsilon_i}\log S_{KL}(x_i,\theta,\epsilon_i) + \lambda_2 R_\theta, \text{where}\\
S_{KL}(x_i,\theta,\epsilon_i)=L_{KL}(x_i,\theta,\epsilon_i)+M_{KL}(x_i,\theta,\epsilon_i)
\label{eq:mat}
\end{aligned}
\end{eqnarray}
$L_{KL}(x_i,\theta,\epsilon_i)$ represents the KL-divergence between the output of adversarial examples and the original examples, $M_{KL}(x_i,\theta,\epsilon_i)$ represents the KL-divergence between the latent representations of adversarial and original examples. $L_{KL}(x_i,\theta,\epsilon_i)$ is defined as the data smoothness in output space and  $M_{KL}(x_i,\theta,\epsilon_i)$ is defined as a the data smoothness in manifold space. 

\begin{figure}[htbp]
	\centering
    \subfigure[Original]{
		\includegraphics[width=0.15\textwidth]{{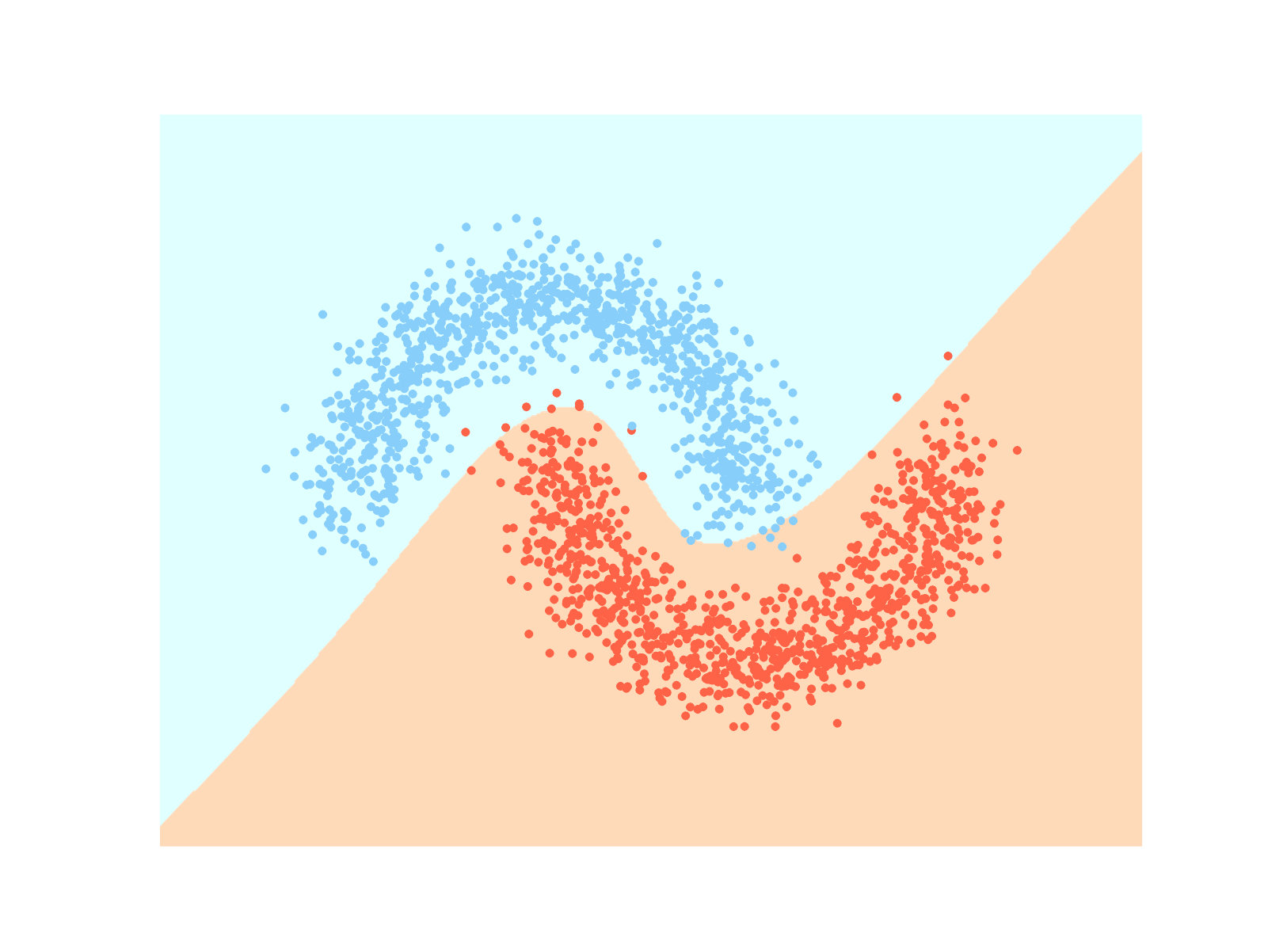}}
		\label{subfig:toy1_org}
		}
	\subfigure[PGD Data]{
		\includegraphics[width=0.15\textwidth]{{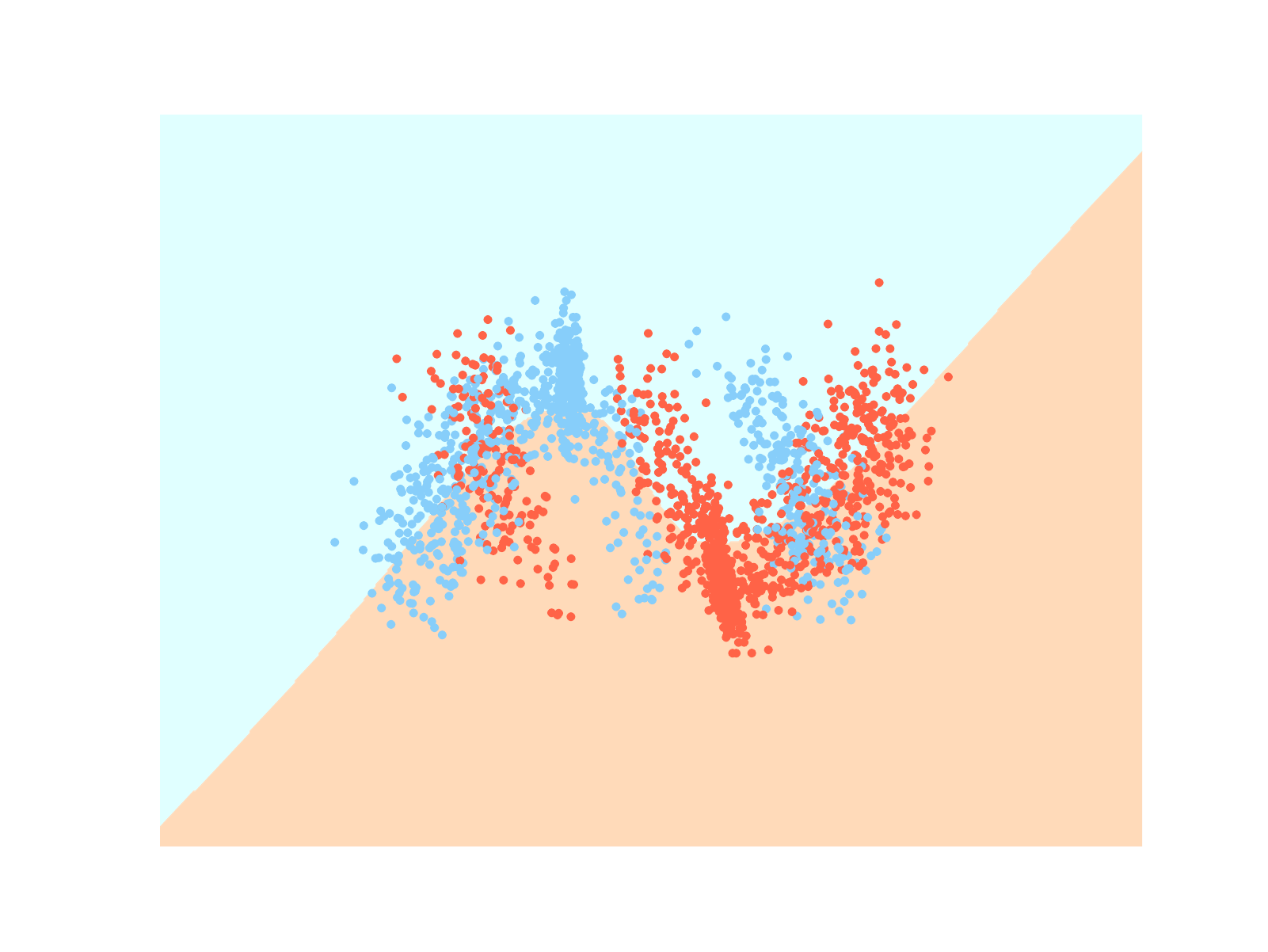}}
		\label{subfig:toy1_pgd}
	}
	\subfigure[FS Data]{
		\includegraphics[width=0.15\textwidth]{{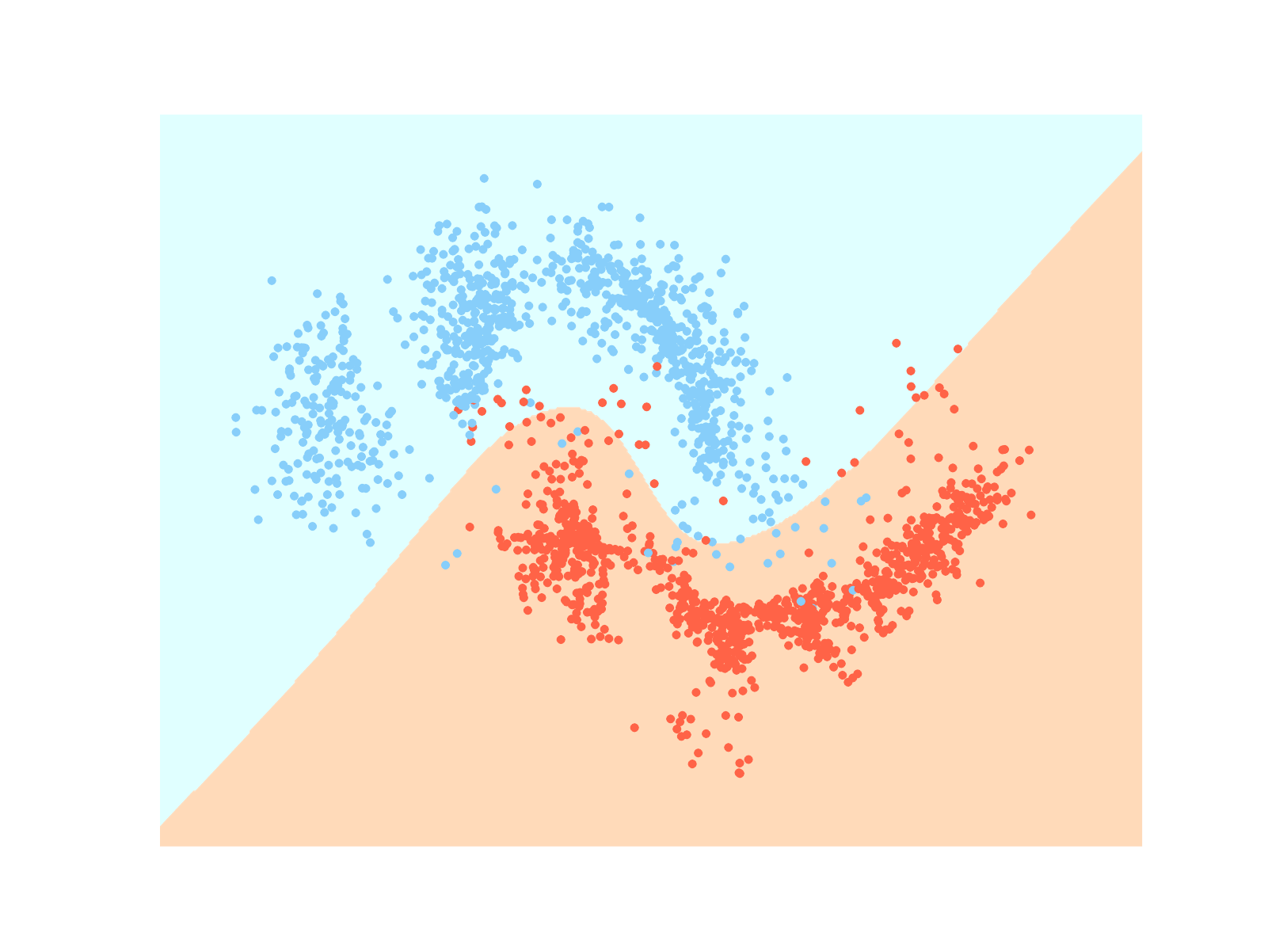}}
		\label{subfig:toy1_fs}
	}
	\subfigure[ATLD Data]{
		\includegraphics[width=0.15\textwidth]{{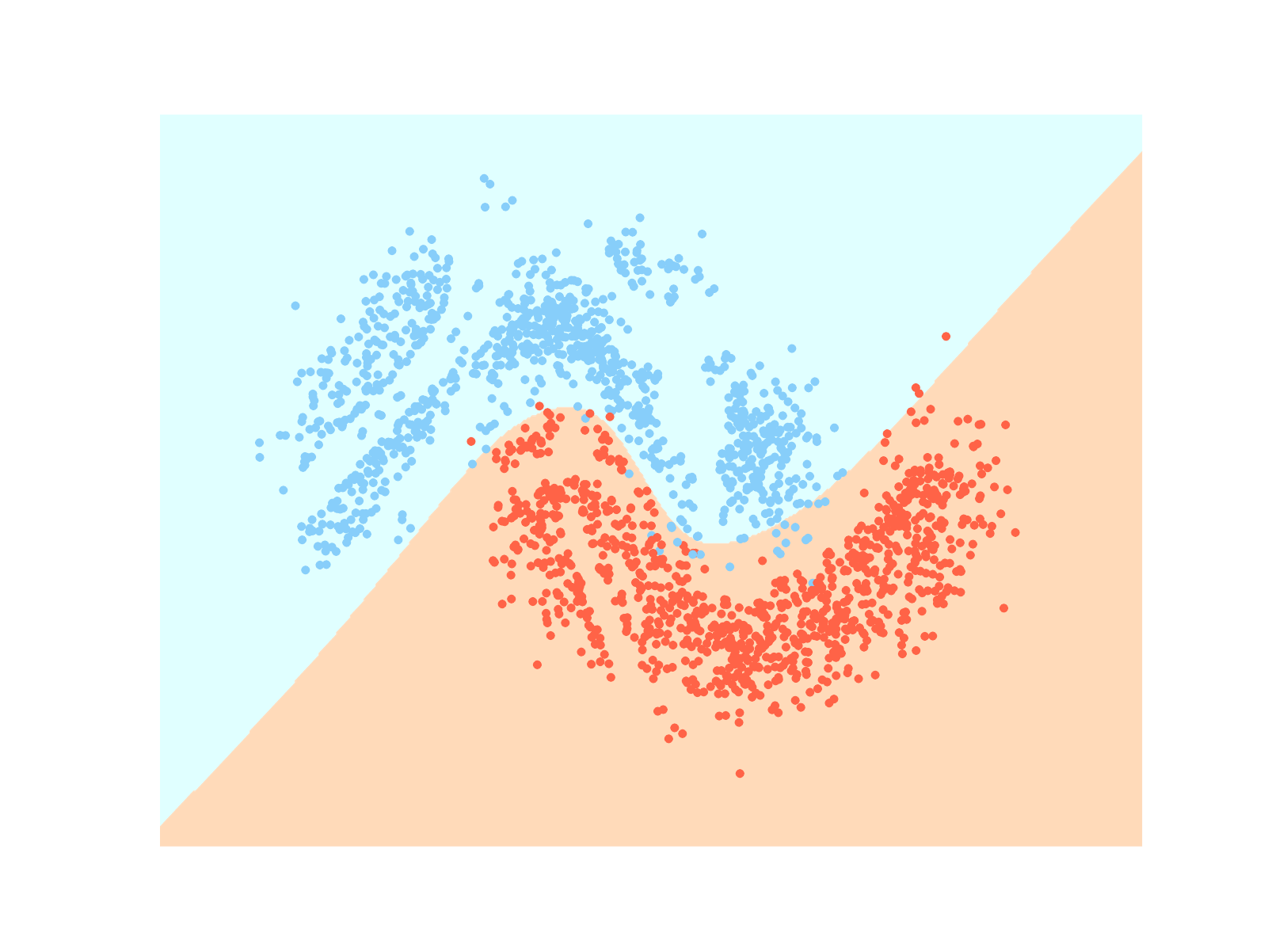}}
		\label{subfig:toy1_ours}
	}\\ \vspace{-5pt}
			\subfigure[Original]{
		\includegraphics[width=0.15\textwidth]{{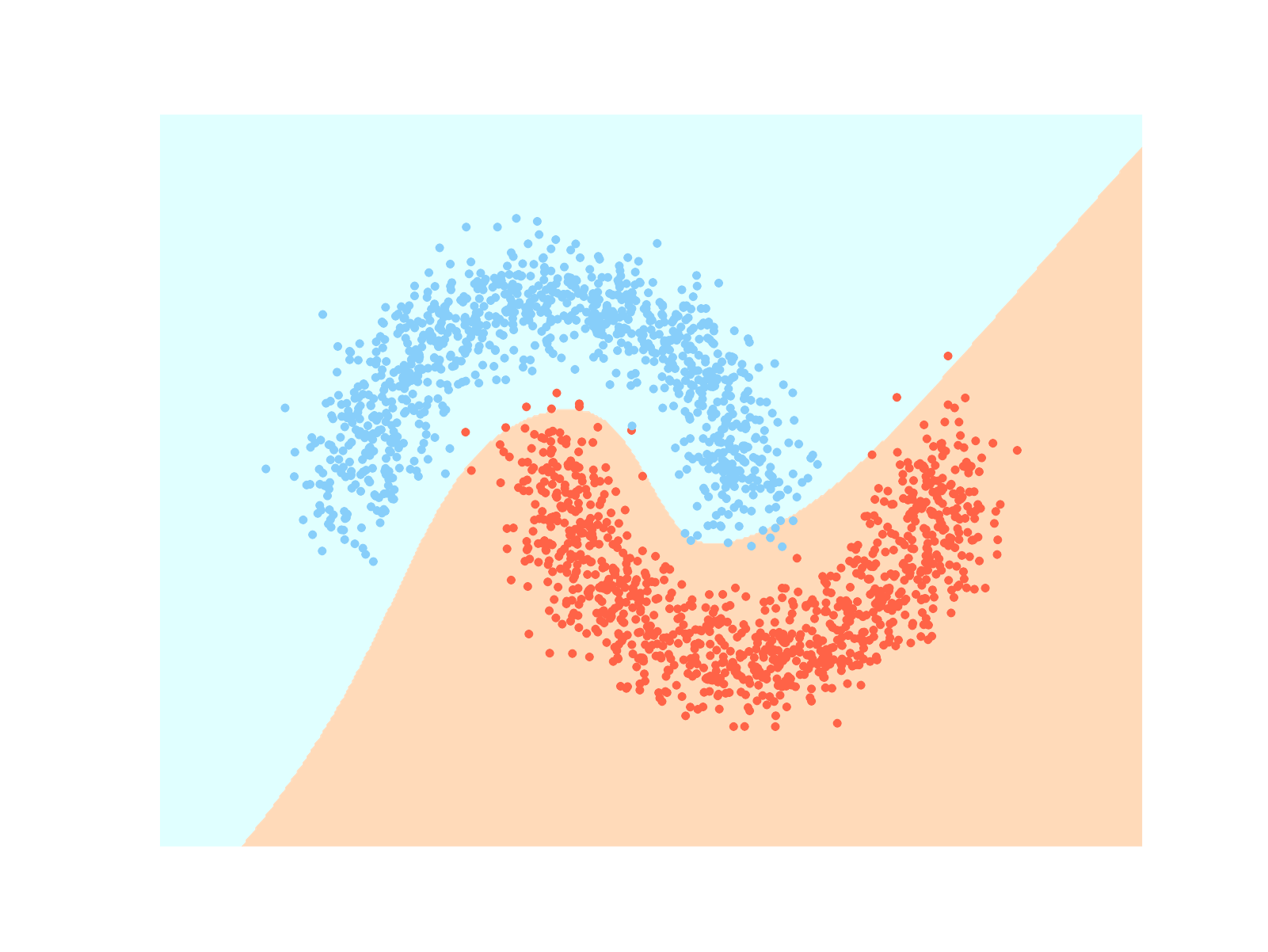}}
		\label{subfig:toy3_org}
	}
	\subfigure[PGD-AT ]{
		\includegraphics[width=0.15\textwidth]{{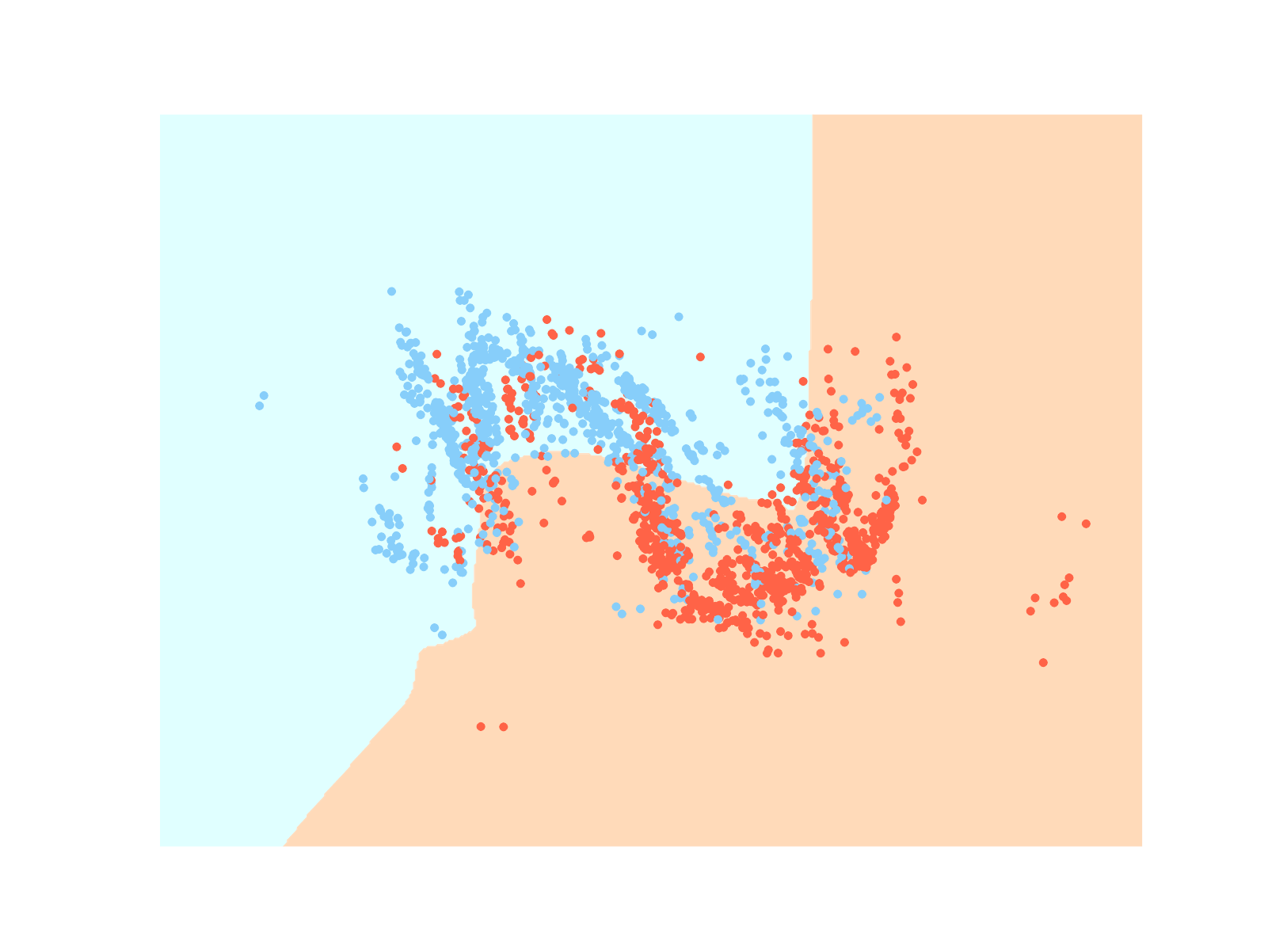}}
		\label{subfig:toy3_pgd}
	}
	\subfigure[FS-AT]{
		\includegraphics[width=0.15\textwidth]{{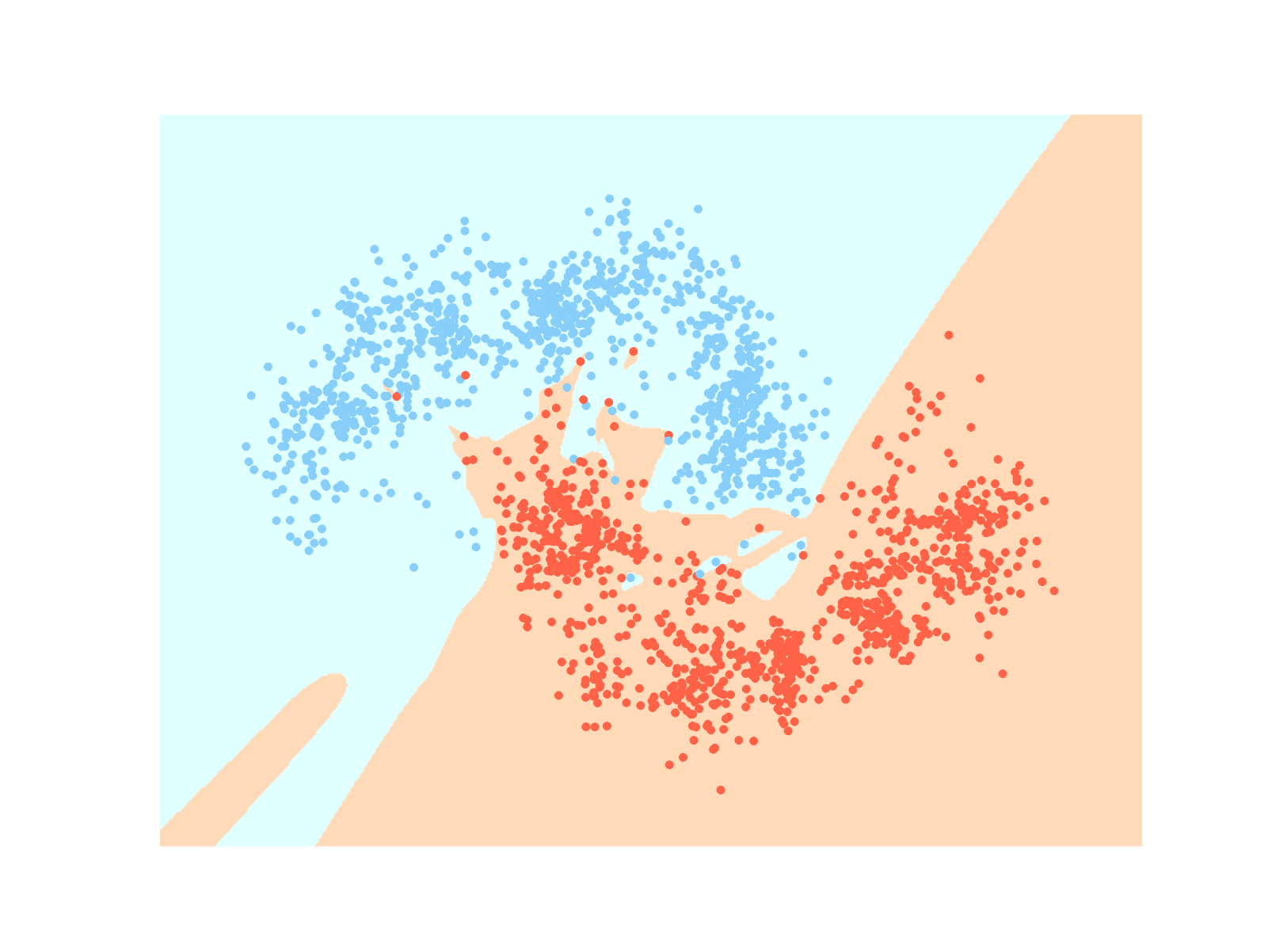}}
		\label{subfig:toy3_fs}
	}
	\subfigure[ATLD]{
		\includegraphics[width=0.15\textwidth]{{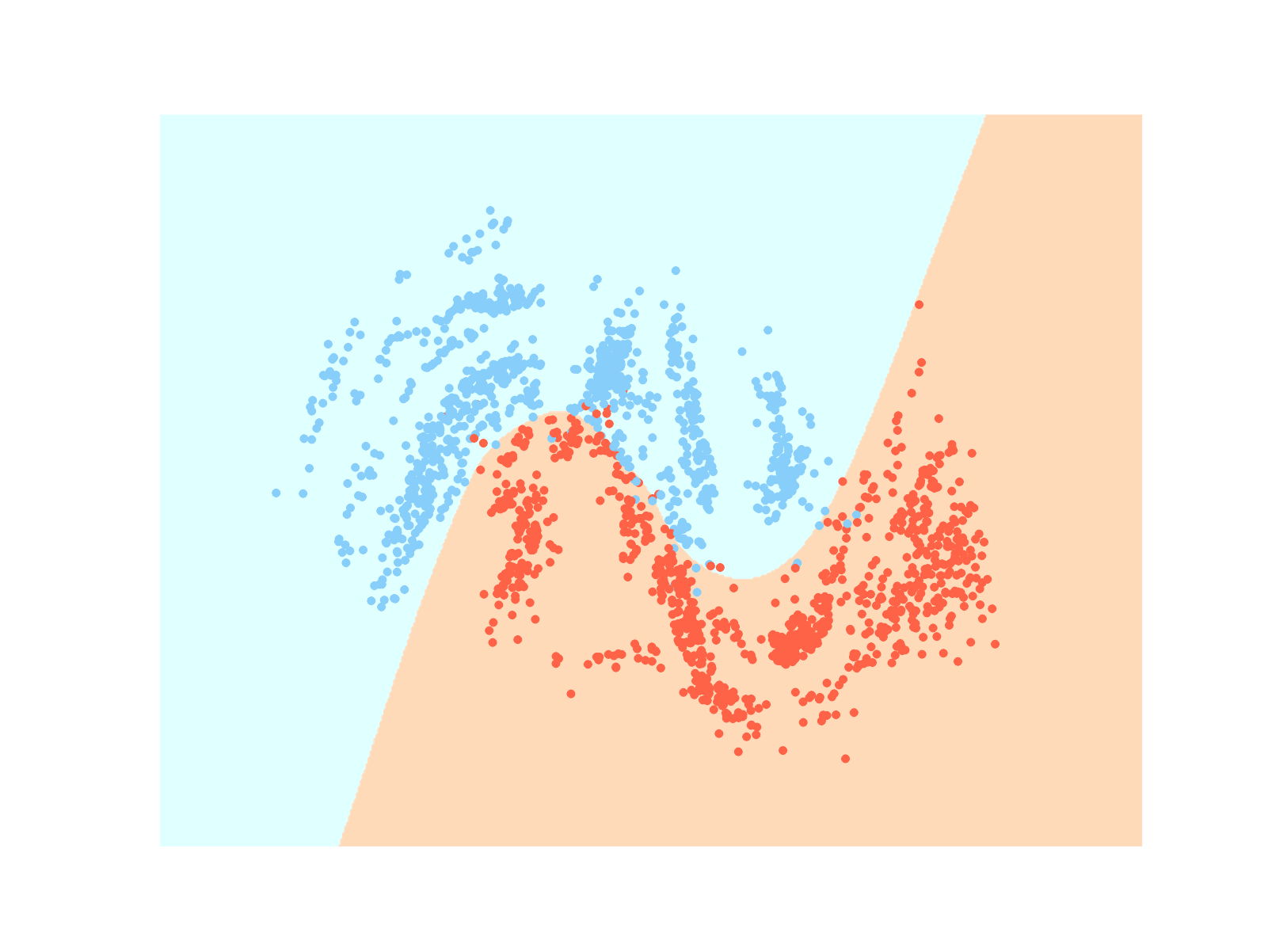}}
		\label{subfig:toy3_ours}
	}

	\caption{ Top: Illustrative different adversarial example; Bottom: Decision boundary after different training. PGD-AT and FS-AT may undermine data manifold~\cite{ATLD}.}	\label{fig:ATLDtoy2}
\end{figure}

\textbf{ATLD}. Similar to FS, Adversarial Training with Latent Distribution (ATLD) argues that  data manifold cannot be ignored when generating adversarial examples~\cite{ATLD}. See Figure~\ref{fig:ATLDtoy2} for illustrations. Different from FS using a fixed metric optimal transport distance that only learns part of the data manifold, ATLD leverages an auxiliary discriminator to learn the data manifold. Owing to the auxiliary Discriminator, a novel Inference method is designed with Manifold Transformation (IMT). IMT aims to push the inference data towards the manifold of natural examples according to the gradient of Discriminator. Specifically every inference data is added with a perturbation which aims to make the perturbed inference data more easier to be classified as natural sample by Discriminator.  Even though ATLD has demonstrated  promising robustness against PGD and C\&W, IMT continues to improve ATLD, even against more sophisticated attacks.

\subsection{Intriguing Properties of Adversarial Training}
\label{sec:proerty_AT}

\textbf{Model Capacity Is Not A bottleneck}. Madry et al.~\cite{PGD} argue that models with larger capacity (e.g., more tunable parameters) will gain more benefits from PGD-AT than models with smaller capacities. They assume that the decision bounds of models should be more complex to separate adversarial examples, therefore requiring more capacity. This argument is proved by the follow-up work~\cite{intriguing2} showing that, unlike traditional classification tasks on ImageNet~\cite{imagenet} that  receives no performance gains from the increased model capacity, the performance of adversarial training can improve accordingly.

\begin{figure}[htbp]
	\centering
	\includegraphics[width=0.4\textwidth]{{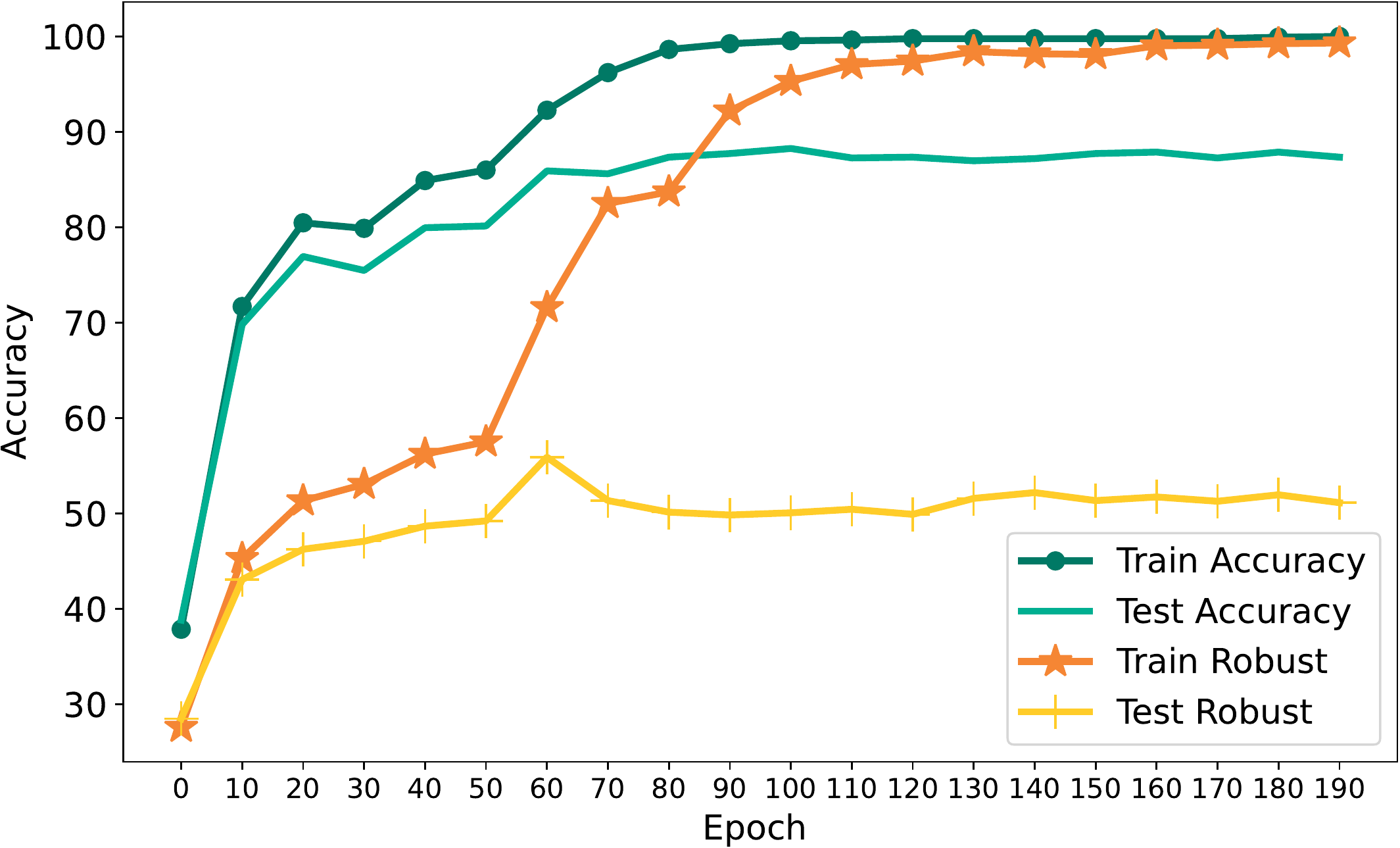}}
	\caption{Learning curves of WideResNet-28-10~\cite{wide_resnet} for PGD-AT on CIFAR-10~\cite{cifar}}. 
	\label{fig:overfitting}
\end{figure}

\textbf{Overfitting in Adversarial Training}. In deep learning, it is a common practice to use over-parameterized networks and to train them as long as possible since such approach does not overly impair the model generalization. However, Rice et al.~\cite{overfitting} indicates that overfitting does occur in adversarial training for deep learning as shown in Figure~\ref{fig:overfitting}. As observed, after the first weight decay at Epoch 60, the test robust accuracy decreases and does not converge to its maximum value, meanwhile the training robust accuracy keeps on increasing. This phenomenon does not occur in the standard accuracy. Motivated from this phenomenon,  an early stop strategy is proposed. Unlike the early stop in FAT~\cite{fat}, the simple strategy uses an earlier checkpoint during training. 


\textbf{Robust Features}. Recent investigations have recently indicated that the PGD-AT trained model tends to utilize features that are strongly correlated with the correct label, effectively reducing the attack surface. Moreover, as argued in~\cite{odds}, standard classifiers utilize too many features that are weakly correlated with the correct label in order to achieve high accuracy, thus sacrificing adversarial robustness. 

Ilyas et al.~\cite{bug} raise the same point. Dividing data features into robust and non-robust ones, they attribute adversarial samples to non-robust features which allow the model to obtain high standard accuracy but low robustness. Models trained on robust features can obtain both high accuracy and robustness. As such, transferability of adversarial sample can also be explained with non-robust features.
To verify the effect of non-robust features, \cite{bug} generated corresponding targeted adversarial samples for each training set (specifying the next class as target mis-classified label for generating adversarial samples), and then trained a DNN with the new generated samples and the real labels. Finally, even though the model was trained with the wrong labels, it still has a 44\% accuracy on the normal test set. The fact that the association between the inputs in the new dataset and their true labels is only maintained by small adversarial perturbations suggests that the adversarial vulnerability is a direct result of the model's sensitivity to features that are better generalized in the data. Furthermore, to validate the effect of robust features, the authors generated a robust training set by feature restriction of the robust model. Empirically, even without adversarial training, training the model directly with this newly generated robust training set alone eventually results in a robust model.

Zhang et al.~\cite{interpreting} show that  adversarially and standard trained models focus on different types of features. They conduct experiments to illustrate the behavior of two types of DNNs on  different transformed data. It is shown that the sensitivity maps based on SmoothGrad~\cite{smoothgrad} of the three models (standard CNN, underfitting CNN and PGD-AT CNN) for different transformed inputs. They conclude that the robust model trained adversarially is more focused on the physical contour information regardless of whether the inputs are transformed, while the standard CNN is more noisy in its focus.  



\section{Further Research Issues}

We turn to discussing some further research issues which have recently drawn much attentions in robust adversarial training, hoping to offer more inspirations and  outlooks in this field. We will review from three aspects: Interpretability of adversarial examples, robust generalization, and robustness evaluation. 

\subsection{Interpretability of Adversarial Examples}
Although great efforts have been made trying to explain adversarial samples, there is no commonly accepted theory. In Section~\ref{sec:visualization}, we have visually explain the cross-model generalization of adversarial examples, it however keeps more fundamental to explain theoretically adversarial samples. Adversarial samples were earlier attributed to the highly nonlinear and inadequate regularization of DNNs. Later, Goodfellow et al.~\cite{AE} argue that they are probably caused due to the linearization of the high-dimensional space. Specifically, denote  $x' = x + \epsilon$ is an adversarial sample, and $x$ is the normal sample. For a linear classifier $f(x) = \theta x+\theta\epsilon$ with parameter $\theta$, even if the infinite norm of $\epsilon$ is very small, due to the large spatial dimensionality, $\theta\epsilon$ could still be relatively large, leading to mis-classifications. A different explanation is given by Szegedy et al.~\cite{intriguing} showing that  DNNs only learn local sub-manifold rather than the entire data manifold due to the limited training samples. However, adversarial samples are mostly distributed in these low probability regions, resulting in mis-classifications. Tsipras et al.~\cite{odds} and Ilyas et al.~\cite{bug} both make a similar point that there exist both robust and non-robust features in the dataset, and non-robust features lead to adversarial samples as introduced in Section~\ref{sec:proerty_AT}. A new viewpoint reveals that  interpolation almost never happens on any high-dimensional (larger than 100) dataset proposed by Balestriero~\cite{balestriero2021learning}, which may provide new explanatory ideas for adversarial examples.

\subsection{Robust Generalization}
Recently, researchers have found that the robustness of the training set is difficult to generalize to the test set, i.e., robust generalization is difficult to be achieved. For example, a model can achieve about 92\% robust accuracy for adversarial samples on the training set after adversarial training, but it can attain only 53\% robust accuracy on the test set~\cite{esposito2020robust, chen2020more}. 

To address this issue, Schmidt et al.~\cite{schmidt2018adversarially} demonstrated that for simple models, the data complexity of adversarial samples is much higher than that of normal samples, making the robust generalization difficult to achieve. To obtain good robust generalization, more samples will be required. Similarly, Zhai et al.~\cite{zhai2019adversarially} propose Generalized VAT (GVAT) showing that the implementation of robust generalizability requires more unlabeled data only. 

Song et al.~\cite{song2019robust} develop Robust Local
Features for Adversarial Training (RLFAT) which leverages Random Block Shuffle (RBS) transformation to improve the robust generalization by forcing the model to focus on local features. They randomly divide adversarial samples into several fragments and shuffled them  horizontally and vertically. Employing the transformed images for adversarial training achieves good result improvement. The argument is, after randomly disrupting the fragments,  the model can focus more on local robust features, which leads to better robust generalization. 

Wu et al.~\cite{wu2020adversarial} design Adversarial Weight Perturbation (AWP) to enhance the robust generalization of the model. Specifically, AWP adds not only the adversarial perturbation into input samples but also into parameters of the model 
as $\min_{w}\max_{v \in \nu}\frac{1}{n}\sum^{n}_{i=1}\max_{\|x'_i - x_i \|_p\le \epsilon} \ell(f_{w+v}(x'_i), y_i)$,
where $w$ is the original parameter, $v$ is the perturbation for $w$ within feasible region $\nu$. In addition, A theory is developed proving that the adversarial weight perturbation indeed causes a tighter upper bound on the robust generalization. 

Zhang et al.~\cite{SCR} argue that the poor robust generalization is due to the significant dispersion in latent representations of training and test data. Thus, they propose Shift Consistency Regularization to force the latent representations of training and test data set to be consistent during training. 

\subsection{Robustness Evaluation}
Proper model robustness evaluation can provide accurate guidance for model selection and optimization. Many researchers believe that the classification accuracy of a model against some specific adversarial samples alone is not sufficient to judge the robustness of the model, since there might also be some potential attacks that break down a model previously considered to be robust~\cite{weng2018evaluating,croce2019provable, wong2018provable,jordan2019provable}, and vice versa~\cite{carlini2018ground, grosse2017statistical}. Some defense approaches argue that they could defend against a particular attack but were later shown to fail with a slight variation of the attack~\cite{bhagoji2017dimensionality,feinman2017detecting}. On the other hand, existing adversarial methods sometimes 
take different network structures, different parameters, for the attacks, even under different attack settings, making it difficult to  compare various methods fairly. To this end,  attempts are made to standardize   robustness evaluation on some sophisticated attacks, e.g. AA~\cite{AA}. Yet, it keeps still unclear which attacks should be included and why some attacks are better than the others.


Alternatively, Weng et al.~\cite{weng2018evaluating} propose Cross Lipschitz Extreme Value for nEtwork Robustness (CLEVER) criterion to evaluate the robustness of the model. It is proved that the CLEVER score is an upper bound of the minimum radius on $l_p$ norm for generating an adversarial sample. Its physical meaning refers to the minimum specific radius within which no adversarial samples can attack the model. Moreover, the larger the CLEVER score, the larger the minimum attack radius and the more robust it is. Therefore CLEVER can be used as an evaluation criterion for robustness which is independent of the type of attack. Croce et al.~\cite{croce2019provable} also propose the robustness evaluation criterion MMR Universal for all $l_p$ norm ($p > 1$) perturbations. They prove that this criterion achieves a tighter upper and lower bounds on robustness for 1, 2 and infinite norm perturbations. Several researchers have also gone further to conduct robustness evaluation for neural networks with RELU activation functions~\cite{wong2018provable,jordan2019provable}.

\section{Conclusion}
This paper presents a timely and comprehensive survey on robust adversarial training. Starting from fundamentals with definition and notations, we introduce a unified theory which can offer an overall framework that may better understand adversarial training. Visualization, and interpretations are discussed followed by a comprehensive summarization and review of different methodologies. We also address the three important research focus in adversarial training: interpretability, robust generalization, and robustness evaluation, which can stimulate future inspirations as well as research outlook. We hope that readers could gain some understanding and interest in robust adversarial training, both theoretically and experimentally, as a way to expand the research community in pattern recognition.


\begin{sidewaystable}[htbp]
\caption{Taxonomy of adversarial attacks covered in this paper.}
\scalebox{1}
{
\begin{tabular}{|c|c|c|c|}
\hline
                            & Attacks       & Type                                 & Remarks                                                                                                                                                    \\ \hline
\multirow{13}{*}{White-box} & L-BFGS~\cite{intriguing}        & \multirow{4}{*}{Optimisation-based}  & Early attack using a box-constrained optimization method                                                                                                   \\ \cline{2-2} \cline{4-4} 
                            & AMDR~\cite{AMDR}          &                                      & Close distance between the input and the target-class input in latent space                                                                                \\ \cline{2-2} \cline{4-4} 
                            & DeepFool~\cite{deepfool}      &                                      & Estimate the minimum distance between inputs and decsion boundary                                                                                          \\ \cline{2-2} \cline{4-4} 
                            & C\&W~\cite{CW}          &                                      & \begin{tabular}[c]{@{}c@{}}Powerful empirically-chosen loss function to approximate, \\ an optimisation problem similar to L-BFGS\end{tabular}               \\ \cline{2-4} 
                            & FGSM~\cite{AE}          & \multirow{5}{*}{Gradient-based}      & Find perturbations direction fast with gradient ascent                                                                                                     \\ \cline{2-2} \cline{4-4} 
                            & BIM~\cite{BIM}           &                                      & Multi-step variants of FGSM                                                                                                                                \\ \cline{2-2} \cline{4-4} 
                            & MI-FGSM~\cite{MBIM}       &                                      & BIM with Momentum, faster to converge                                                                                                                      \\ \cline{2-2} \cline{4-4} 
                            & R+FGSM~\cite{R+FGSM}        &                                      & Randomized initialized FGSM which helps to escape from local optimum                                                                                             \\ \cline{2-2} \cline{4-4} 
                            & PGD~\cite{PGD}           &                                      & Randomized initialized BIM, powerful attack to evaluate robustness                                                                                         \\ \cline{2-4} 
                            & BPDA~\cite{BPDA}          & \multirow{4}{*}{Approximation-based} & \begin{tabular}[c]{@{}c@{}}Replace non-differentiable parts with differentiable parts, \\ to overcome gradient masking\end{tabular}                         \\ \cline{2-2} \cline{4-4} 
                            & SPSA~\cite{SPSA}          &                                      & gradient estimation method to overcome gradient masking                                                                                                    \\ \cline{2-2} \cline{4-4} 
                            & ATN~\cite{ATN}           &                                      & Generate adversarial perturbation with neural networks                                                                                                     \\ \cline{2-2} \cline{4-4} 
                            & AdvGAN~\cite{advgan}        &                                      & Generate adversarial perturbation with GAN                                                                                                                 \\ \hline
\multirow{6}{*}{Black-box}  & SBA~\cite{SBA}           &                                      & Generate transfer attack on a substitute model that imitates the target model                                                                              \\ \cline{2-4} 
                            & Zoo~\cite{ZOO}           &                                      & \begin{tabular}[c]{@{}c@{}}Approximate the gradients of the objective function \\ using finite-difference numerical estimates similar to SPSA\end{tabular} \\ \cline{2-4} 
                            & OPA~\cite{OPA}           &                                      & Deceive the target model by only one pixel                                                                                                                  \\ \cline{2-4} 
                            & BA~\cite{boundary_attack}            &                                      & \begin{tabular}[c]{@{}c@{}}Start from one existing adversarial sample and \\ randomly walk to search the decision boundary\end{tabular}                      \\ \cline{2-4} 
                            & NAA~\cite{NAA}           &                                      & \begin{tabular}[c]{@{}c@{}}Generate natural adversarial examples with a pre-trained generative model \\ and a learnable generative model\end{tabular}      \\ \cline{2-4} 
                            & Square Attack~\cite{andriushchenko2020square} &                                      & \begin{tabular}[c]{@{}c@{}}Randomly search the score-based\\ black-box attack for norm bounded perturbations\end{tabular}                                    \\ \hline
\multirow{2}{*}{More Sophisticated}          & AutoAttack~\cite{AA}    &                                      & Ensemble white/black-box attacks, and target/non-target attacks                                                                                         \\\cline{2-4}
                            & Rays~\cite{Rays}          &                                      & Use binary search to locate the decision boundary radius                                                                                                   \\ \hline
\end{tabular}
}
\label{tab:attack}
\end{sidewaystable}


\begin{sidewaystable}[htbp]
\caption{Taxonomy of adversarial training covered in this paper.}
\begin{tabular}{|c|c|c|}
\hline
                                                                                  & Defence     & Remark                                                                                                                                                            \\ \hline
\multirow{14}{*}{Conventional AT}                                                 & FGSM-AT~\cite{AE}     & Train a model with FSGM adversarial examples                                                                                                                        \\ \cline{2-3} 
                                                                                  & PGD-AT~\cite{PGD}      & Train a model with PGD examples, import baseline                                                                                                                    \\ \cline{2-3} 
                                                                                  & TLA~\cite{TLA}         & Train a model with the triplet loss                                                                                                                                 \\ \cline{2-3} 
                                                                                  & ANL~\cite{adversarial_noise_layer}         & Inject adversarial perturbation into latent features                                                                                                               \\ \cline{2-3} 
                                                                                  & BAT~\cite{bilateral}   & Perturb both the image and the label during training                                                                                                              \\ \cline{2-3} 
                                                                                  & EAT~\cite{ensemble}         & Train a model with adversarial examples generated on other pre-trained models                                                                                        \\ \cline{2-3} 
                                                                                  & PED~\cite{ensemble_diversity}         & Force non-maximal predictions  as diverse as possible in an ensemble system                                       \\ \cline{2-3} 
                                                                                  & ALP~\cite{ALP}         & Force adversarial examples and their corresponding natural samples to have similar output                                     \\ \cline{2-3} 
                                                                                  & FAT~\cite{fat}         & Train a model with friendly adversarial examples which do not cross the decision boundary too much                     \\ \cline{2-3} 
                                                                                  & Overfitting~\cite{overfitting} & leverage early stop to choose the best checkpoint to inference           \\ \cline{2-3}
                                                                                  &
                                              Free~\cite{free} & Accelerate AT by recycling the gradient information 
                                                                    \\ \cline{2-3}
                                                                                  &
                                              SLAT~\cite{park2021reliably} & Accelerate AT with the single-step latent adversarial training (SLAT) \\ \cline{2-3}
                                                                                  &
                                              GradAlign~\cite{andriushchenko2020understanding} & Accelerate AT with GradAlign that aims to maximize the gradient alignment between $x$ and $x'$\\ \cline{2-3}
                                                                                  &
                                              Fast AT~\cite{wong2020fast} &  Accelerate AT with R+FGSM~\cite{R+FGSM}, Cyclic learning rate~\cite{cyclical}, and Mixed-precision arithmetic~\cite{micikevicius2017mixed}
                                                                                  \\ \hline
\multirow{5}{*}{Manifold AT}                                                      & TRADES~\cite{trades}      & Trade-off the relationship between correctness and robustness                                                                                                     \\ \cline{2-3} 
                                                                                  & VAT~\cite{virtual_adversarial_training}         & Train a model with virtual adversarial examples                                                                                                                     \\ \cline{2-3} 
                                                                                  & FS~\cite{fs}          & \begin{tabular}[c]{@{}c@{}}Maximising the optimal transport (OT) distance between\\ natural examples and perturbed examples to scatter features\end{tabular}    \\ \cline{2-3} 
                                                                                  & MAT~\cite{zhang2021manifold}         & Train the model with considering smoothness in output and latent space                                                                                                \\ \cline{2-3}
                                                                                  & ATLD~\cite{ATLD}        & \begin{tabular}[c]{@{}c@{}}Train a model with adversarial examples which perturb\\ the latent manifold most through a auxiliary discriminaotr\end{tabular}        \\ \hline
\multirow{4}{*}{\begin{tabular}[c]{@{}c@{}}Robust \\ Generalization\end{tabular}} & GVAT~\cite{zhai2019adversarially}        & Train model with more adversarial examples generated from unlabel data                                                                                            \\ \cline{2-3} 
                                                                                  & RLFAT~\cite{song2019robust}       & \begin{tabular}[c]{@{}c@{}}Train a  model with an adversarial example which is divided into\\ patches and randomly shuffled in horizontal and vertical \end{tabular}\\ \cline{2-3} 
                                                                                  & AWP~\cite{wu2020adversarial}         & Train a model with perturbation parameter of model and adversarial examples                                                                                         \\ \cline{2-3} 
                                                                                  & SCR~\cite{SCR}         & Force the latent representations of training and test data set to be consistent                                       \\ \hline
\end{tabular}
\label{tab:defence}
\end{sidewaystable}

\bibliography{mybibfile}

\end{document}